\documentclass[preprint]{elsarticle}

\usepackage{amsmath}
\usepackage{graphicx}
\usepackage{epstopdf}
\usepackage{lineno,hyperref}
\usepackage{soul}
\usepackage{bm}
\usepackage{mathtools}

\usepackage{xcolor}
\definecolor{vorange}{RGB}{255, 160, 0}
\definecolor{sugrey}{RGB}{200, 200, 200}
\definecolor{cogreen}{RGB}{50, 220, 30}

\newcommand{\norm}[1]{\left\lVert #1\right\lVert}
\newcommand{\bvare}{\bm{\varepsilon}}
\newcommand{\es}[2]{\bvare_{#1\gets #2}}
\newcommand{\jac}[2]{\frac{\partial\mathbf{s}_{#1}}{\partial\mathbf{s}_{#2}}}

\newcommand{\stau}[1]{\mathbf{s}_{#1}}
\newcommand{\stt}{\stau{t}}
\newcommand{\stp}{\stau{t-1}}
\newcommand{\xtau}[1]{\mathbf{x}_{#1}}
\newcommand{\xtt}{\xtau{t}}
\newcommand{\sbart}{\bar{\mathbf{s}}_t}
\newcommand{\gv}[1]{\mathcal{G}_{#1_t}}

\DeclarePairedDelimiter{\set}{\{}{\}}
\DeclarePairedDelimiter{\floor}{\lfloor}{\rfloor}

\makeatletter
\def\ps@pprintTitle{%
   \let\@oddhead\@empty
   \let\@evenhead\@empty
   \def\@oddfoot{\small\textit{Preprint}}
   \let\@evenfoot\@oddfoot
}
\makeatother

\begin{document}

\begin{frontmatter}

\title{Learning Longer-term Dependencies via Grouped Distributor Unit}

\author[mainaddress]{Wei Luo}
\ead{willi4m@zju.edu.cn}

\author[mainaddress]{Feng Yu\corref{corrauthor}}
\cortext[corrauthor]{Corresponding author}

\address[mainaddress]{College of Biomedical Engineering and Instrument Science, 
Yuquan campus, Zhejiang University, 38 Zheda Road, Hangzhou 310027, China}

\begin{abstract}
  Learning long-term dependencies still remains difficult for recurrent neural networks (RNNs)
  despite their success in sequence modeling recently.
  In this paper, we propose a novel gated RNN structure, which contains only one gate.
  Hidden states in the proposed grouped distributor unit (GDU) are partitioned into groups. 
  For each group, the proportion of memory to be overwritten in each state transition is limited to a constant
  and is adaptively distributed to each group member.
  In other word, every separate group has a fixed overall update rate, yet all units are allowed to have different paces.
  Information is therefore forced to be latched in a flexible way, which helps the model to capture long-term dependencies in data.
  Besides having a simpler structure, GDU is demonstrated experimentally to outperform LSTM and GRU on 
  tasks including both pathological problems and natural data set.
\end{abstract}

\begin{keyword}
Recurrent neural network\sep Sequence learning\sep Long-term memory
\end{keyword}

\end{frontmatter}

\section{Introduction}

Recurrent Neural Networks (RNNs, \cite{rumelhart86, werbos88}) are powerful dynamic systems for tasks that involve sequential inputs, 
such as audio classification, machine translation and speech generation.
As they process a sequence one element at a time, internal states are maintained to store information computed from the past inputs which
makes RNNs capable of modeling temporal correlations between elements from any distance in theory.

In practice, however, it is difficult for RNNs to learn long-term dependencies in data by using back-propagation through time 
(BPTT, \cite{rumelhart86}) due to the well known \emph{vanishing} and \emph{exploding gradient} problem \cite{hochreiter01}.
Besides, training RNNs suffers from gradient conflicts (e.g. \emph{input conflict} and \emph{output conflict} \cite{lstm})
which make it challenging to latch long-term information while keeping mid- and short-term memory simultaneously.
Various attempts have been made to increase the temporal range that credit assignment takes effect for recurrent models during training,
including adopting a much more sophisticated Hessian-Free optimization method instead of stochastic gradient descent 
\cite{martens10, martens11}, using \emph{orthogonal} weight matrices to assist optimization \cite{saxe13, irnn} and allowing direct 
connections to model inputs or states from the distant past \cite{narx, dilatedrnn, mistrnn}.
Long short-term memory (LSTM, \cite{lstm}) and its variant, known as gated recurrent units (GRU, \cite{gru}) 
mitigate gradient conflicts by using \emph{multiplicative gate units}.
Moreover, the vanishing gradient problem is alleviated by the additivity in their state transition operator.
Simplified gated units have been proposed \cite{ugrnn, mgu} yet the ability of capturing long-term dependencies has not been improved.
Recent work also supports the idea of partitioning the hidden units in an RNN into separate modules with different processing periods
\cite{cwrnn}.

In this paper, we introduce Grouped Distributor Unit (GDU), a new gated recurrent architecture with additive state transition and
only one gate unit.
Hidden states inside a GDU are partitioned into groups, each of which keeps a constant proportion of previous memory at each time step,
forcing information to be latched.
The vanishing gradient problem, together with the issue of gradient conflict, which impede the extraction of long-term dependencies
are thus alleviated.

We empirically evaluated the proposed model against LSTM and GRU on both synthetic problems which are designed 
to be pathologically difficult and natural dataset containing long-term components.
Results reveal that our proposed model outperforms LSTM and GRU on these tasks with a simpler structure and less parameters.

\section{Background and related work}\label{ss:bk}

An RNN is able to encode sequences of arbitrary length into a fixed-length representation by folding a new observation 
$\mathbf{x}_t$ into its hidden state $\mathbf{s}_t$ using a transition operator $T$ at each time step $t$:
\begin{equation}
  \mathbf{s}_t = T(\mathbf{x}_t, \mathbf{s}_{t-1})
  \footnote{We do not consider RNNs with connections from the past such as NARX RNN \cite{narx}.}
\end{equation}
Simple recurrent networks (SRN, \cite{elman90}), known as one of the earliest variants, make $T$ as the composition of an 
element-wise nonlinearity with an affine transformation of both $\mathbf{x}_t$ and $\mathbf{s}_{t-1}$:
\newcommand{\netst}{\mathbf{W}_s\mathbf{x}_t + \mathbf{U}_s\mathbf{s}_{t-1} + \mathbf{b}_s}
\begin{equation}
  \mathbf{s}_t = \phi_s(\netst)
\end{equation}
where $\mathbf{W}_s$ is the input-to-state weight matrix, $\mathbf{U}_s$ is the state-to-state recurrent weight matrix, 
$\mathbf{b_s}$ is the bias and $\phi_s$ is the nonlinear activation function.
For the convenience of the following descriptions, we denote this kind of operators as $\eta(\cdot, \cdot, \phi)$,
and a subscript can be added to distinguish different network components.
Thus in SRN, $\mathbf{s}_t = \eta_s(\mathbf{x}_t, \mathbf{s}_{t-1}, \phi_s)$.

\newcommand{\esfv}[2]{\frac{\partial\mathcal{L}_{#1}}{\partial \mathbf{s}_{#2}}}
\newcommand{\esfc}[1]{\frac{\partial\mathcal{L}_t}{\partial s_\tau^{#1}}}
During training via BPTT, the error obtained from the output of an RNN at time step t (denoted as $\mathcal{L}_t$)
travels backward through each state unit. 
The corresponding error signal propagated back to time step $\tau$ 
(denoted as $\es{\tau}{t}=\esfv{t}{\tau}$\footnote{
  $\esfv{t}{\tau} = (\esfc{1}, \esfc{1}, \cdots, \esfc{M})^T$,
  in which $M$ is the state size and the $k$-th component $\esfc{k}$ represents the sensitivity of 
  $\mathcal{L}_t$ to small perturbations in the $k$-th state unit at time step $\tau$. 
},
$\tau<t$)
contains a product of $t-\tau$ Jacobian matrices:
\begin{equation}
  \es{\tau}{t} = \es{t}{t}\prod_{t\ge i>\tau}\jac{i}{i-1} 
  \label{eq_chain}
\end{equation}
From Eq. (\ref{eq_chain}) we can easily find a \emph{sufficient condition} for the \emph{vanishing gradient} problem to occur, 
i.e. $\forall \tau<i\le t, \norm{\jac{i}{i-1}} < 1$.
Under this condition, a bound $\xi \in \mathcal{R}$ can be found such that $\forall i, \norm{\jac{i}{i-1}} \le\xi < 1$, and
\begin{equation}
  \norm{\es{\tau}{t}} = \norm{\es{t}{t}\prod_{t\ge i>\tau}\jac{i}{i-1}} \le \xi^{t-\tau}\norm{\es{t}{t}}
\end{equation}
As $\xi < 1$, long term contributions (for which $t-\tau$ is large) go to $0$ exponentially fast with $t - \tau$.

In SRN, $\jac{i}{i-1}$ is given by $\mathbf{U}_s^T diag({\phi'_s(\netst)})$.
As a result, if the derivative of the nonlinear function is bounded in SRN, namely,
$\exists\kappa \in \mathcal{R}$, s.t. $|\phi'_s(x)| \le \kappa$, it will be \emph{sufficient} for $\lambda_1 < \frac{1}{\kappa}$,
where $\lambda_1$ is the largest singular value of the recurrent weight matrix $\mathbf{U}_s$,
for $\es{\tau}t$ to vanish (as $t\to \infty$)\cite{pascanu13}.

Any RNN architecture with a long-term memory ability should at least be designed to make sure the norm of its transition Jacobian  
will not easily be bounded by $1$ for a long time span as it goes through a sequence.

\subsection{Gated additive state transition (GAST)}

Long short-term memory (LSTM, \cite{lstm}) introduced a memory unit with self-connected structure which can maintain its
state over time, and non-linear gating units (originally input and output gates) which control the information flow into and out of it.
Since the initial proposal in 1997, many improvements have been made to the LSTM architecture \cite{forgetgate, peephole}.
In this paper, we refer to the variant with forget gate and without peephole connections, which has a comparable performance
with more complex variants \cite{greff17}:
\newcommand{\rn}[3]{\eta_{#1}(\mathbf{x}_t, \mathbf{#2}_{t-1}, #3)}
\begin{subequations}
\begin{align}
  \mathbf{f}_t & = \rn{f}{h}{\sigma} \\
  \mathbf{i}_t & = \rn{i}{h}{\sigma} \\ 
  \mathbf{o}_t & = \rn{o}{h}{\sigma} \\
  \mathbf{\bar{\mathbf{s}}}_t & = \rn{\bar{s}}{h}{\mathrm{tanh}} \\
  \mathbf{s}_t & = \mathbf{f}_t \odot \mathbf{s}_{t-1} + \mathbf{i}_t \odot \bar{\mathbf{s}}_t \\
  \mathbf{h}_t & = \mathbf{o}_t \odot \mathrm{tanh}(\mathbf{s}_t)
\end{align}
\end{subequations}
Here $\sigma$ denotes the sigmoid activation and $\odot$ denotes element-wise multiplication. 
Note that $\mathbf{h}_t$ should also be considered as hidden state besides $\stt$.

Cho et al. \cite{gru} proposed a similar architecture with gating units called gated recurrent unit (GRU). 
Different from LSTM, GRU exposes all its states to the output and use a linear interpolation between the previous state 
$\mathbf{s}_{t-1}$ and the candidate state $\bar{\mathbf{s}}_t$:
\begin{subequations}
\begin{align}
  \mathbf{r}_t & = \rn{r}{s}{\sigma} \\
  \mathbf{z}_t & = \rn{z}{s}{\sigma} \\
  \mathbf{\bar{\mathbf{s}}}_t & = \eta_{\bar{s}}(\mathbf{x}_t, \mathbf{r}_t\odot\mathbf{s}_{t-1}, \mathrm{tanh}) \\
  \mathbf{s}_t & = \mathbf{z}_t \odot \mathbf{s}_{t-1} + (1 - \mathbf{z}_t) \odot \bar{\mathbf{s}}_t 
\end{align}
\end{subequations}

Previous work has clearly indicated the advantages of the gating units over the more traditional recurrent units \cite{chung14}.
Both LSTM and GRU perform well in tasks that require capturing long-term dependencies. 
However, the choice of these two structures may depend heavily on the dataset and corresponding task.

\begin{figure}[htpb]
  \centering
  \includegraphics[width=\textwidth,height=\textheight,keepaspectratio]{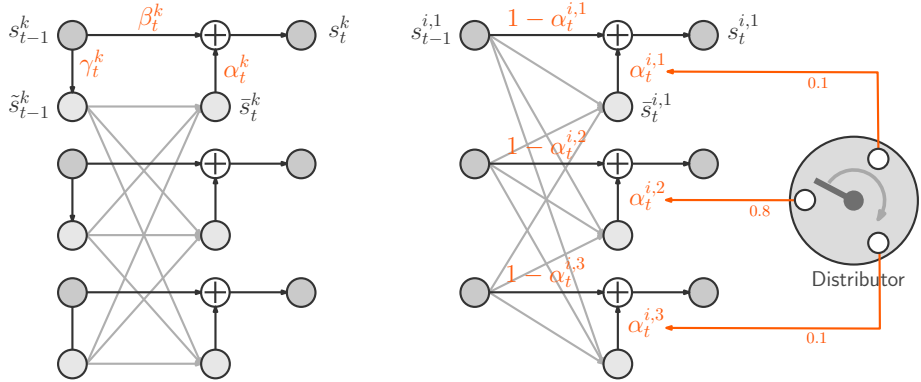} 
  \caption{
    \textbf{Left}:
    The gated additive state transition (GAST).
    Inputs and outputs are not shown.
    Superscript $k$ denotes the ordinal number of a state unit.
    In LSTM, $\tilde{s}_{t-1}^k$ corresponds to $h_{t-1}^k$.
    In GRU, $\beta_t^k = 1 - \alpha_t^k$.
    \textbf{Right}:
    The GAST in a GDU group with size $3$ and $\delta_i=1.0$.
    Compared to LSTM and GRU, gate operator $\gamma_t$ is removed and gate operators $\{\alpha_t^{i,k}\}_k$ inside the group
    is correlated, i.e. $\sum_{k}\gv{\alpha}^{i,k} = \delta_i = 1$.
    Any unit assigned with a high $\gv{\alpha}^{i,k}$ will force other group members to latch information.
   }
  \label{gast_gdu}
\end{figure}

It is easy to notice that the most prominent feature shared between these units is the additivity in their state transition operators.
In another word, both LSTM and GRU keep the existing states and add the new states on top of it 
instead of replacing previous states directly, as it did in traditional recurrent units like SRN.
Another important ingredient in their transition operator is the gating mechanism,
which regulates the information flow and enables the network to form skip connections adaptively.
In this paper we refer to this kind of transition operators as the \emph{Gated Additive State Transition} (GAST) with a general formula:
\newcommand{\gate}[2]{#1_t(#2)}
\begin{subequations}\label{eqn:general}
\begin{align}
  \sbart   & = \eta_{\bar{s}}(\xtau{t}, \gate{\gamma}{\stau{t-1}}, \phi) \\
  \stau{t} & = \gate{\beta}{\stau{t-1}} + \gate{\alpha}{\sbart} \label{eq_gast}
\end{align}
\end{subequations}
where $\alpha_t$, $\beta_t$ and $\gamma_t$ are called \emph{gate operators} with subscript $t$ indicating that values of the 
corresponding gating units change over time (see Fig. \ref{gast_gdu} (left)). 
In LSTM:
\begin{subequations}
\begin{align}
  \gate{\gamma}{\stau{t-1}} & = \mathbf{o}_{t-1} \odot \mathrm{tanh}(\stau{t-1}) \label{equ:lstmgamma} \\ 
  \gate{\beta}{\stau{t-1}}  & = \mathbf{f}_t \odot \stau{t-1} \\ 
  \gate{\alpha}{\sbart}     & =\mathbf{i}_t \odot \sbart
\end{align}
\end{subequations}
whilst in GRU:
\begin{subequations}
\begin{align}
  \gate{\gamma}{\stau{t-1}} & = \mathbf{r}_{t} \odot \stau{t-1} \\ 
  \gate{\beta}{\stau{t-1}}  & = \mathbf{z}_t \odot \stau{t-1} \\ 
  \gate{\alpha}{\sbart}     & = (1 - \mathbf{z}_t) \odot \sbart
\end{align}
\end{subequations}
We denote the gate vector used in a gate operator $T$ at time step $t$ as $\gv{T}$.
Note that except Eq. (\ref{equ:lstmgamma}), gate operators $T_t$ have a common form 
\footnote{In the following part of this paper, \emph{gate operators} are referred to as being in this form.}:
\begin{equation}
  T_t(\mathbf{s}) = \gv{T} \odot \mathbf{s}
\end{equation}
where $\mathbf{s}$ is a state vector to be gated.
We use $\beta_t = 1 - \alpha_t$ to indicate $\gv{\beta} = 1 - \gv{\alpha}$ as in the case of GRU.
According to Eq. (\ref{eq_gast}), the transition Jacobian of a GAST can be resolved into 4 parts:
\begin{equation}
  \jac{t}{t-1} = J_{\stau{t-1}} + J_{\sbart} + J_{\gv{\alpha}} + J_{\gv{\beta}}
\end{equation}
in which
\begin{subequations} \label{jstp}
\begin{align}
  J_{\stau{t-1}}  & = diag(\gv{\beta}) \label{eqn:j1} \\
  J_{\sbart}      & = \frac{\partial\sbart}{\partial\stau{t-1}}\cdot diag(\gv{\alpha}) \label{eqn:j2} \\
  J_{\gv{\beta}}  & = \frac{\partial\gv{\beta}}{\partial\stp}\cdot diag(\stp) \label{eqn:j3} \\
  J_{\gv{\alpha}} & = \frac{\partial\gv{\alpha}}{\partial\stp}\cdot diag(\sbart) \label{eqn:j4}
\end{align}
\end{subequations}
The gradient property of GAST is much better than that of SRN since it can easily prevent its transition Jacobian norm to be 
bounded within $1$ by saturating part of units in $\gv{\beta}$ nearly at 1.
Intuitively, when this happens, the corresponding components of error signal are allowed to be back-propagated easily 
through the shortcut created by the additive character of GAST without vanishing too quickly.

The original LSTM \cite{lstm} uses \emph{full gate recurrence} \cite{odyssey}, 
which means that all neurons receive recurrent inputs from all gate activations at the previous time step 
besides the block outputs. 
Nevertheless, it still follows Eqs. (\ref{eqn:general}).
Another difference is that the original LSTM does not use forget gate, 
i.e. $\beta_t(\stp) = \stp$,
thus in Eq. (\ref{eqn:j1}), $J_{\stau{t-1}}$ is a unit diagonal matrix $\mathcal{I}$.
In addition, gradients are truncated by replacing the other components in its transition Jacobian,
i.e. Eqs. (\ref{eqn:j2}), (\ref{eqn:j3}) and (\ref{eqn:j4}), by zero, forming a \emph{constant error carrousel} (CEC) 
where $\jac{t}{t-1} = \mathcal{I}$.
It is noticeable, however, that if the gradients are not truncated,
Eq. (\ref{eq_chain}) does not hold for LSTMs since the gate vector $\mathbf{o}_{t-1}$
used in $\gamma_t$ is calculated at the previous time step, see Eq. (\ref{equ:lstmgamma}).
In this condition, a concatenation of $\stt$ and $\mathbf{h}_t = \gamma_t(\stt)$ should be used in analysis of its transition Jacobian, 
as in Fig. \ref{fig:merg_norm}.

Simplifying GAST has drawn interest of researchers recently.
GRU itself reduces the gate units to $2$ compared to LSTM which has $3$ gate units by 
coupling forget gate and input gate into one update gate, namely making the gate operator $\beta_t$ equals to $1 - \alpha_t$.
In this paper we denote this kind of GAST as cGAST, with the prefix c short for \emph{coupled}.
Based on GRU, the Minimal Gated Unit (MGU, \cite{mgu}) reduced the gate number further to only 1 by letting 
$\gamma_t = \beta_t = 1 - \alpha_t$ without losing GRU's accuracy benefits.
The Update Gate RNN (UGRNN, \cite{ugrnn}) entirely removed $\gamma_t$ operator. 
However, none of these models has shown superiority over LSTM and GRU on long-term tasks with single-layer hidden states.

\subsection{Units partitioning}

Although the capacity of capturing long-term dependencies in sequences is of crucial importance of RNNs,
it is worthwhile to notice that the flowing data is usually embedded with both slow-moving and fast-moving information,
of which the former corresponds to long-term dependencies.
Along with the existence of both long- and short-term information in sequences,
the training process always has \emph{gradient conflict} existing.
Here \emph{gradient conflict} mainly refers to the contradiction between error signals back-propagated to a same time step, but
injected at different time steps during training via BPTT.
This issue may hinder the establish of long-term memory even without the gradient vanishing problem.

\newcommand{\esk}[2]{\varepsilon^k_{#1\gets #2}}
Consider a task in which a GRU is given one data point at a time and assigned to predict the next, e.g. $m$ERG (see Section \ref{sss:merg}). 
If the correct prediction at time step $t_1$ is heavily depending on the data point appeared at time step $t_0$, namely $x_{t_0}$, 
where $t_0 \ll t_1$, we can say a long-term dependency exists between $x_{t_0}$ and $x_{t_1+1}$.
GRU can capture this kind of dependency by learning to encode $x_{t_0}$ into some state units and latch it until $t_1$. 
For simplicity, let us focus on a single state unit $s^k$ and assume that the information of $x_{t_0}$ has been stored in $s_{t_0}^k$.
At time step $t$ ($t_0 < t < t_1$), state unit $s_t^k$ will often receive conflicting error signals.
The error signal $\esk{t}{t_1}$ injected at time step $t_1$ may attempt to make $s_t^k$ keep its value until $t_1$.
While other error signals injected before $t_1$, say, $t_2$, may hope that $s_{t_2}^k$ helps to do the prediction at time step $t_2$,
thus it may attempt to make $s_t^k$ to be overwritten by a new value.
This conflict makes the GRU model hesitate to shut the update gate for $s^k$ by setting $\gv{\alpha}^k$ to $0$.
In GRU, we also observed that state units latching long-term memories 
(with corresponding neurons in $\gv{\beta}$ staying active for a long time) are usually sparse (see Fig. \ref{fig:gxu_on_10erg} (left)),
which impedes the back-propagation of effective long-term error signals, since short-term error signals dominate.
As a result, learning can be slow.

El Hihi and Bengio first showed that RNNs can learn both long- and short-term dependencies more easily and efficiently 
if state units are partitioned into groups with different timescales \cite{hihi95}. The clockwork RNN (CW-RNN) \cite{cwrnn} 
implemented this by assigning each state unit a fixed temporal granularity, making state transition happens only at its
prescribed clock rate.
It can also be seen as a member of cGAST family.
More specifically, a UGRNN with a special gate operator $\beta_t$ in which 
each gate vector value $\gv{\beta}^k$ is explicitly scheduled to saturate at either $0$ or $1$.
CW-RNN does not suffer from gradient conflict for it inherently has the ability to latch information.
However, the clock rate schedule should be tuned for each task.

\section{Grouped Distributor Unit}

\newcommand{\pu}{\mathcal{P}_{\alpha_t}}
\newcommand{\pk}{\mathcal{P}_{\beta_t}}
As introduced in Section \ref{ss:bk}, a network combining the advantages of GAST and 
the idea to partition state units into groups seems promising.
Further, we argue that a dynamic system with memory does not need to overwrite the vast majority of its memory 
based on relatively little input data.
For cGAST models whose $\beta_t = 1 - \alpha_t$, we define the proportion of states to be overwritten at time step $t$ as:
\begin{equation}
  \pu = \frac{1}{K}\sum_{k=1}^K \gv{\alpha}^k
\end{equation}
where $K$ is the state size. 
On the other hand, the proportion of previous states to be kept is:
\begin{equation}
  \pk = \frac{1}{K}\sum_{k=1}^K \gv{\beta}^k = 1 - \pu
\end{equation}
Hence in our view, if a model input $\xtt$ contains small amount of information compared to system memory $\stp$, 
$\pu$ should be kept low to protect the previous states.
For cGAST family members, a lower $\pu$ leads to more active units in $\gv{\beta}$ (see Fig.\ref{fig:gxu_on_10erg} (right)) and thus 
less prone to be affected by gradient conflict.

\newcommand{\gs}[3]{s_{#3}^{#1,#2}}
To put a limit on $\pu$, we start by a plain UGRNN and partition its state units into $N$ groups:
\begin{equation}
  \stt = \left\{\left\{s_t^{i,j}\right\}_{j=1}^{M_i}\right\}_{i=1}^N
\end{equation}
where the $i$-th group contains $M_i$ units. 
At each time step, for each $i$, we let a positive constant $\delta_i < M_i$ to be \textbf{distributed} to the 
corresponding components in $\gv{\alpha}$, namely 
\begin{equation}
  \sum_{j=1}^{M_i} \gv{\alpha}^{i,j} = \delta_i, ~i = 1, 2, \cdots, N
\end{equation}
Thus $\pu$ becomes a constant given by
\begin{equation}
  \pu = \frac{\sum_{i=1}^{N}\delta_i}{\sum_{i=1}^N M_i} = \frac{1}{K}\sum_{i=1}^N\delta_i \in (0, 1)
\end{equation}
See Fig.\ref{gast_gdu} (right), the distribution work in each group is done by a \emph{distributor},
hence the proposed structure is called \emph{Grouped Distributor Unit} (GDU).
The distributor is implemented by utilizing the softmax activation over each group individually in calculating $\gv{\alpha}$:
\newcommand{\nta}[1]{\vartheta_t^{#1}}
\newcommand{\dtij}{d_t^{i,j}}
\begin{subequations} \label{eqn:sog}
\begin{align}
  \bm{\vartheta}_t  & = \mathbf{W}_\alpha\mathbf{x}_t + \mathbf{U}_\alpha\mathbf{s}_{t-1} + \mathbf{b}_\alpha \\
  \dtij             & = \frac{\exp(\nta{i,j})}{\sum_{j=1}^{M_i}\exp(\nta{i,j})} \\
  \gv{\alpha}^{i,j} & = 
  \left\{ 
    \begin{array}{ll}
      \delta_i\cdot\dtij   & \textrm{if $\delta_i\in(0,1]$} \\
      \frac{M_i-\delta_i}{M_i-1}\cdot\dtij + \frac{\delta_i-1}{M_i-1}   & \textrm{if $\delta_i\in(1,M_i)$} 
    \end{array}
  \right.
\end{align}
\end{subequations}
here $1\le i\le N,~1\le j\le M_i$ and $\bm{\vartheta}_t = (\nta{1,1},\cdots,\nta{1,M_1},\cdots,\nta{N,1},\cdots,\nta{N,M_N})^T$.
\footnote{The permutation of $\left\{\left\{\vartheta_t^{i,j}\right\}_{j=1}^{M_i}\right\}_{i=1}^N$ can be arbitrary.}
Note that $\gv{\alpha}^{i,j} \in [0,\delta_i)$ when $\delta_i\in(0,1]$ 
and $\gv{\alpha}^{i,j} \in (\frac{\delta_i-1}{M_i-1}, 1]$ when $\delta_i \in (1, M_i)$.
The resulting GDU is given by
\begin{subequations}
\begin{align}
  \mathbf{a}_t & = \zeta(\mathbf{W}_\alpha\mathbf{x}_t + \mathbf{U}_\alpha\mathbf{s}_{t-1} + \mathbf{b}_\alpha; 
  \{\delta_i,M_i\}_{i=1}^N) \\
  \stt         & = (1 - \mathbf{a}_t)\odot\stp + \mathbf{a}_t\odot 
  \mathrm{tanh}(\mathbf{W}_s \mathbf{x}_t + \mathbf{U}_s \stp + \mathbf{b}_s)
\end{align}
\end{subequations}
where $\zeta(\cdot, \{\delta_i,M_i\}_{i=1}^N) $ denotes distributor operator with group configuration $\{\delta_i,M_i\}_{i=1}^N$
as is detailed in Eqs. (\ref{eqn:sog}).

In this paper, we let $\delta_i=1, i=1,2,\cdots,N$.  
As a consequence,
\begin{equation}
  \pu = \frac{N}{\sum_{i=1}^N M_i} = \frac{N}{K}
\end{equation}
If the size of each state group is set to a constant $M$, $\pu$ will be further reduced to $\frac{1}{M}$.

GDU has an inherent strength to keep a long-term memory since any saturated state unit $s^{i,j}$  
will force all other group members to latch information.
As a result, ``bandwidth'' is wider for long-term information to travel forward and error signals to back-propagate 
(see Fig.\ref{fig:gxu_on_10erg} (right)). 

Like CW-RNN, we set an explicit rate $\delta_i$ for each group.
However, instead of making all group members act in the same way, we allow each unit to find its own rate by learning.

\section{Experiments}

We evaluated the proposed GDU on both pathological synthetic tasks and natural data set in comparison with LSTM and GRU.
It is important to point out that although LSTM and GRU have similar performance in nature data set \cite{chung14}, 
one model may outperform another by a huge gap in different pathological tasks like the adding problem (see \ref{sss:ap})
at which GRU is good and the temporal order problem (see \ref{sss:to}) in which LSTM performs better.

If not otherwise specified, all networks have one hidden layer with a same state size.
Weight variables were initialized via Xavier uniform initializer \cite{glorot10}, and the initial values of all internal state
variable were set to $0$.
All networks were trained using Adam optimization method \cite{adam} via BPTT, and the models were implemented using Tensorflow \cite{tf}. 
In GDU models, $\delta_i = 1$ apply to all groups.

\subsection{The adding problem}\label{sss:ap}

The adding problem is a sequence regression problem which was originally proposed in \cite{lstm} to examine the ability
of recurrent models to capture long-term dependencies.
Two sequences of length $L$ are taken as input.
The first one consists of real numbers sampled from a uniform distribution in $[0, 1]$.
While the second sequence serves as indicators with exactly two entries being 1 and the remaining being 0.
We followed the settings in \cite{urnn} where $L$ is a constant and the first 1 entry is located uniformly at random
in the first half of the indicator sequence, whilst the second 1 entry is located uniformly at random in another half.
The target of this problem is to add up the two entries in the first sequence whose corresponding indicator in the
second sequence is 1. 
A naive strategy of outputting $1$ regardless of the inputs yields a mean squared error of $0.167$,
which is the variance of the sum of two independent uniform distributions over $[0, 1]$.
We took it as the baseline.

Four different lengths of sequences, $L\in\set{200, 1000, 5000, 10000}$ were used in this experiment.
For each length, $500$ sequences were generated for testing,
while a batch of $20$ sequences were randomly generated at each training step.  
Four models, an LSTM with $100$ hidden states, a GRU with $100$ hidden states, a GDU with $10$ groups of size $10$ and 
a GDU with only $1$ group of size $10$ were compared, with the corresponding parameter number $41.3K$, $31.0K$, $20.7K$ and $271$.
A simple linear layer without activation is stacked on top of the recurrent layer in each model.

\begin{figure}[htpb]
  \centering
  \includegraphics[width=0.49\textwidth,keepaspectratio]{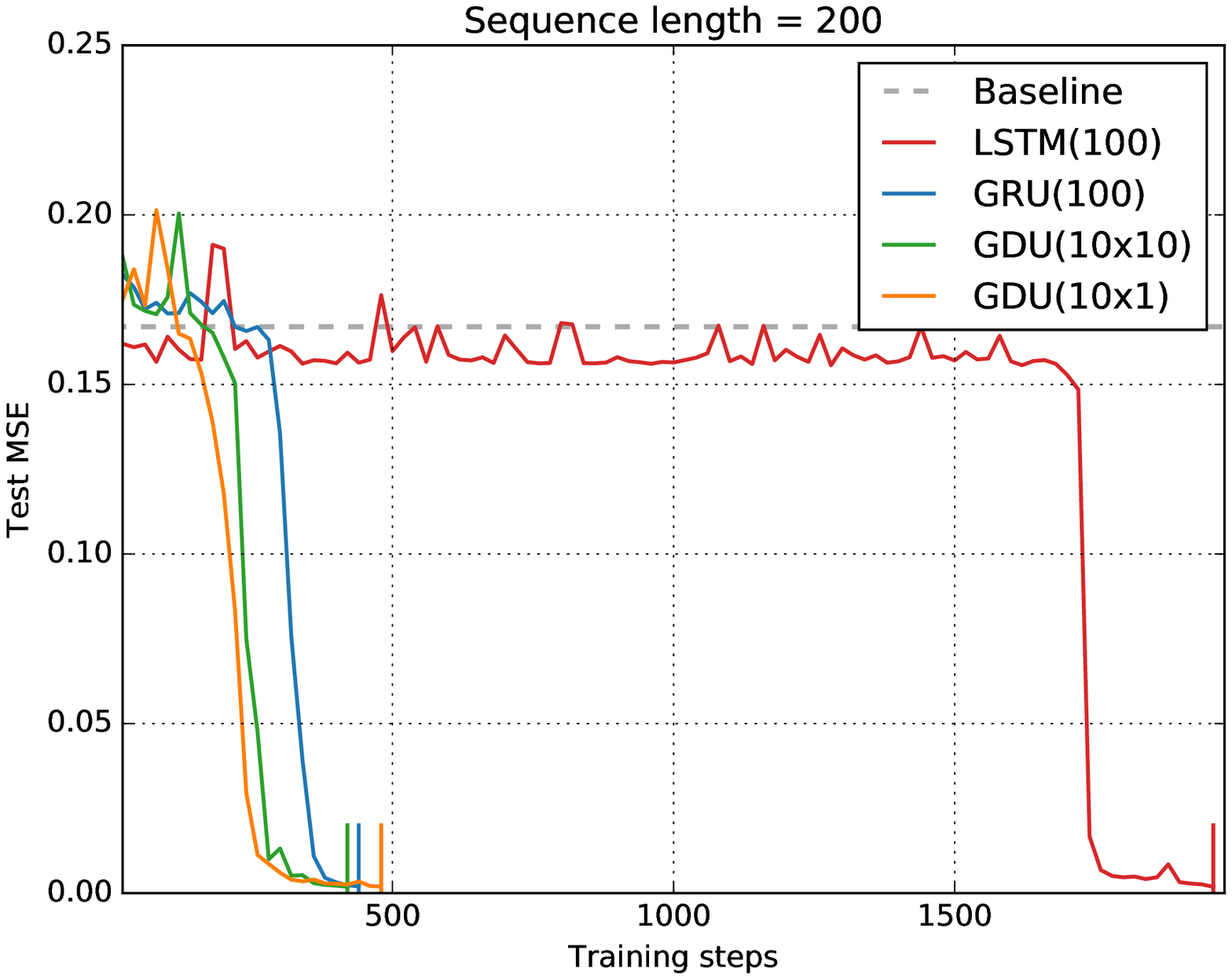}
  \includegraphics[width=0.49\textwidth,keepaspectratio]{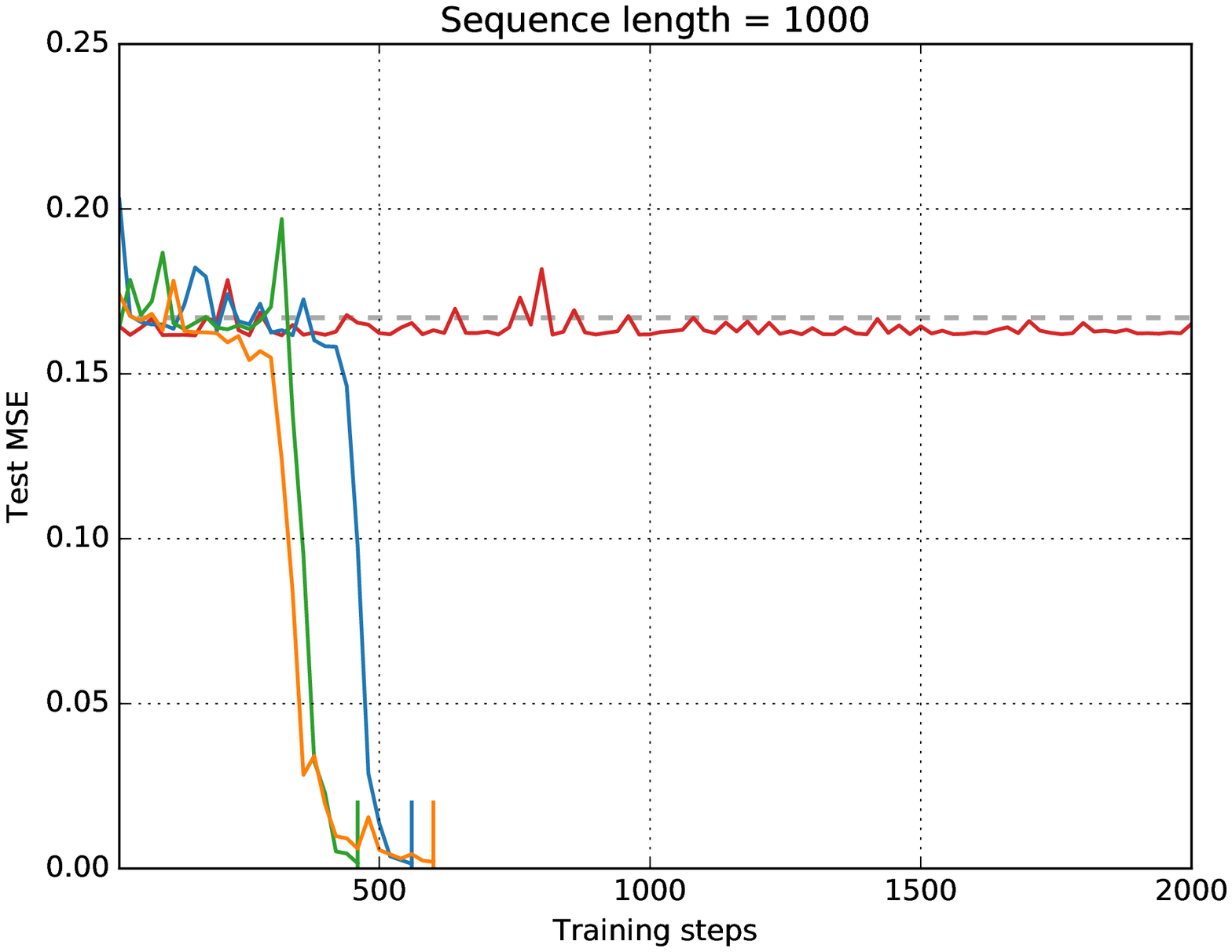}
  \includegraphics[width=0.49\textwidth,keepaspectratio]{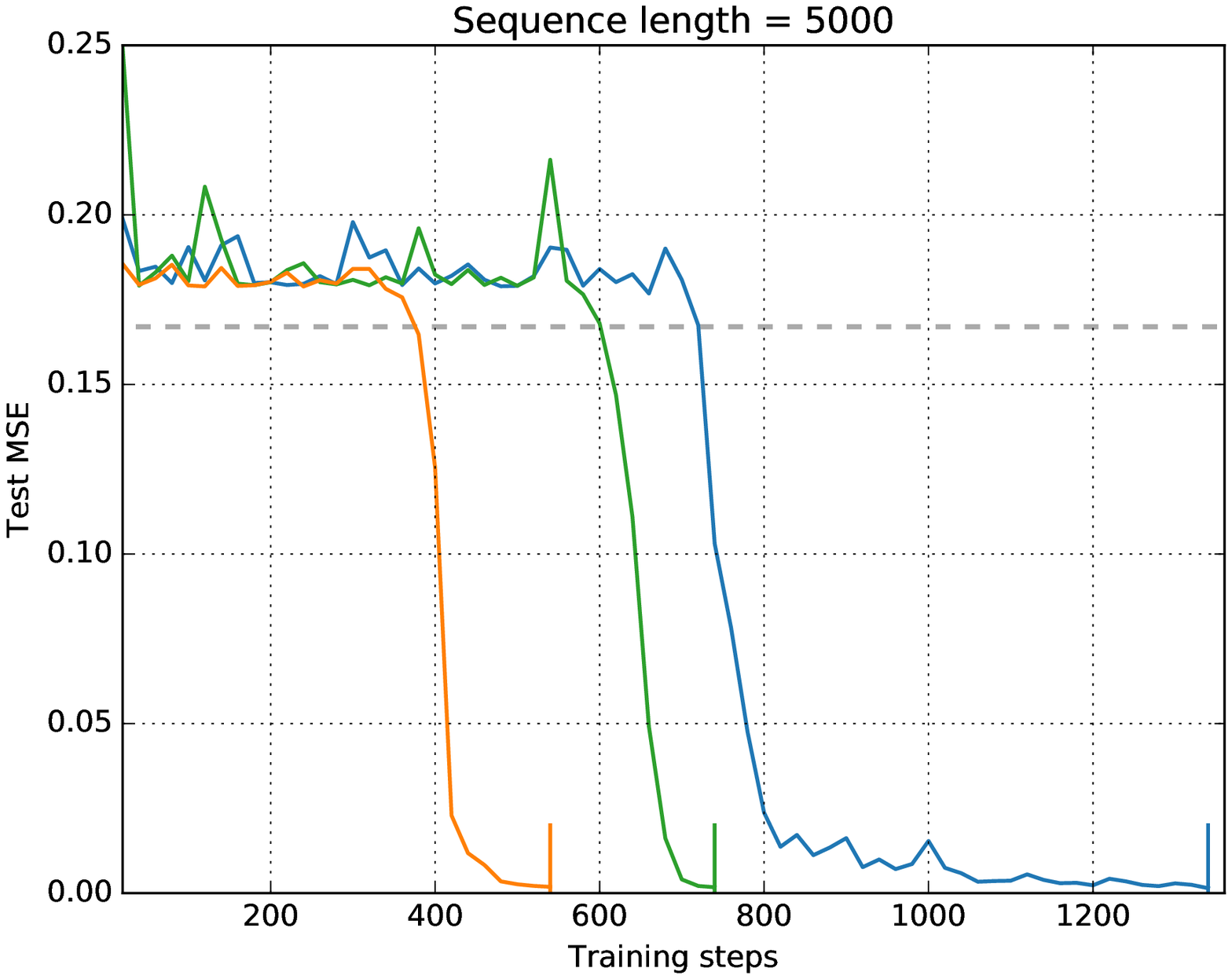}
  \includegraphics[width=0.49\textwidth,keepaspectratio]{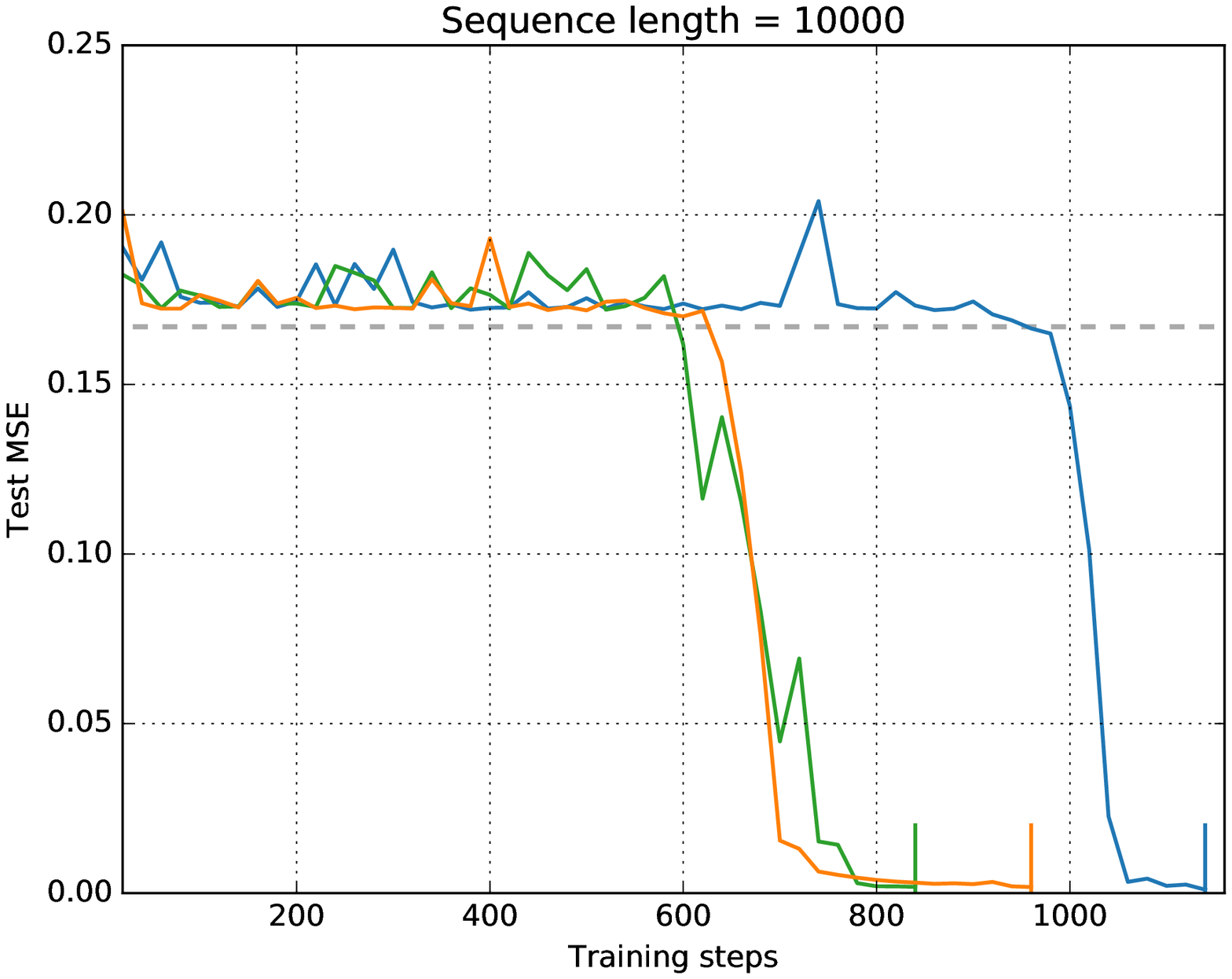}
  \caption{
    The results of the adding problem on different sequence lengths. 
    The legends for all sub-figures are the same thus are only shown in the first sub-figure, 
    in which state sizes are specified following model names.
    For a GDU model, $(M\times N)$ means it has $N$ groups of size $M$.
    Each training trial was stopped when the test MSE reached below $0.002$, as indicated by a short vertical bar.
    When training with sequences of length $1000$, 
    LSTM(100) failed to converge within $10000$ steps and only the curve of the first $2000$ steps is shown.
  }
  \label{fig:ap}
\end{figure}

The results are shown in Fig. \ref{fig:ap}.
Obviously GRU outperforms LSTM in these trials.
LSTM fails to converge within $10000$ training steps when $L$ is $1000$ while GRU can learn this task within $1300$ steps
even trained with sequences of length $10000$.
Our GDU models perform slightly better than GRU with less parameters.
As $L$ increases, this advantage becomes more obvious.
Note that a GDU with only one group of size $10$ has comparable performance with a much bigger one,
which indicates that GDU can efficiently capture simple long-term dependencies even with a tiny model.

\subsection{The 3-bit temporal order problem}\label{sss:to}

The 3-bit temporal order problem is a sequence classification problem to examine the ability of recurrent models
to extract information conveyed by the temporal order of widely separated inputs of recurrent models \cite{lstm}.
The input sequence consists of randomly chosen symbols from the set $\set{a, b, c, d}$ except for three elements at 
position $t_1$, $t_2$ and $t_3$ that are either $X$ or $Y$.
Position $t_k$ is randomly chosen between $\floor{\frac{(k-1)\cdot L}{3}}$ and $\floor{\frac{(k-1)\cdot L}{3}} + 10$,
where $k=1, 2, 3$ and $L$ is the sequence length.
The target is to classify the order (either XXX, XXY, XYX, XYY, YXX, YXY, YYX, YYY)
which is represented locally using $8$ units, as well as the input symbol (represented using $6$ units).

Four different lengths of sequences, $L\in\set{100, 200, 500, 1000}$ were used in this experiment.
Same with the settings in \ref{sss:ap}, we generated $500$ testing sequences for each length,
and randomly generated a batch of $20$ sequences for each training step. 
Accuracy is used as the metric on testing set, and the baseline is $0.125$.
We compared an LSTM model with $100$ hidden states, a GRU model with $100$ hidden states and a GDU with $10$ groups of size $10$
on these data sets.
The parameter numbers are $43.6K$, $32.9K$ and $22.2K$ respectively.

\begin{figure}[htpb]
  \centering
  \includegraphics[width=0.49\textwidth,keepaspectratio]{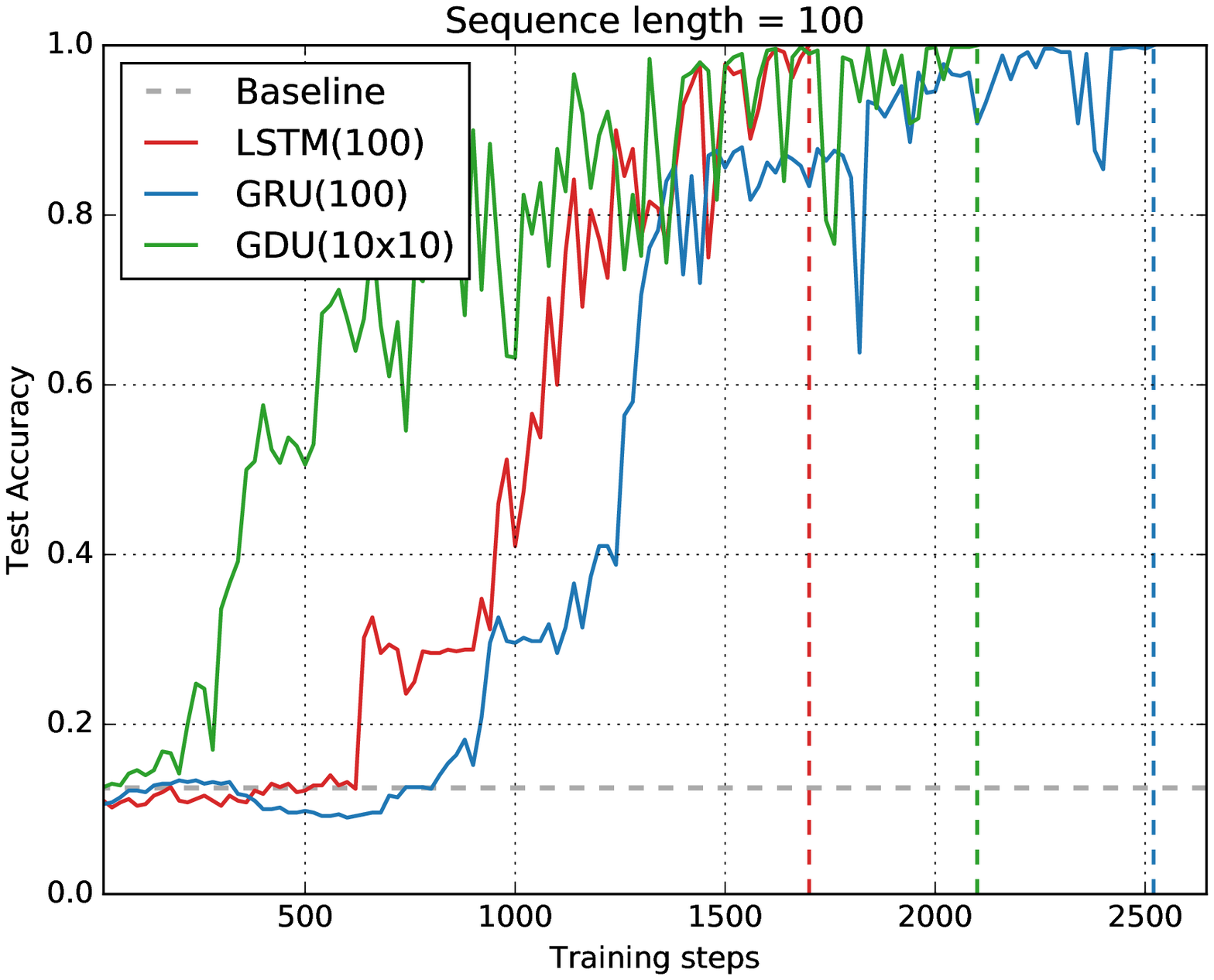}
  \includegraphics[width=0.49\textwidth,keepaspectratio]{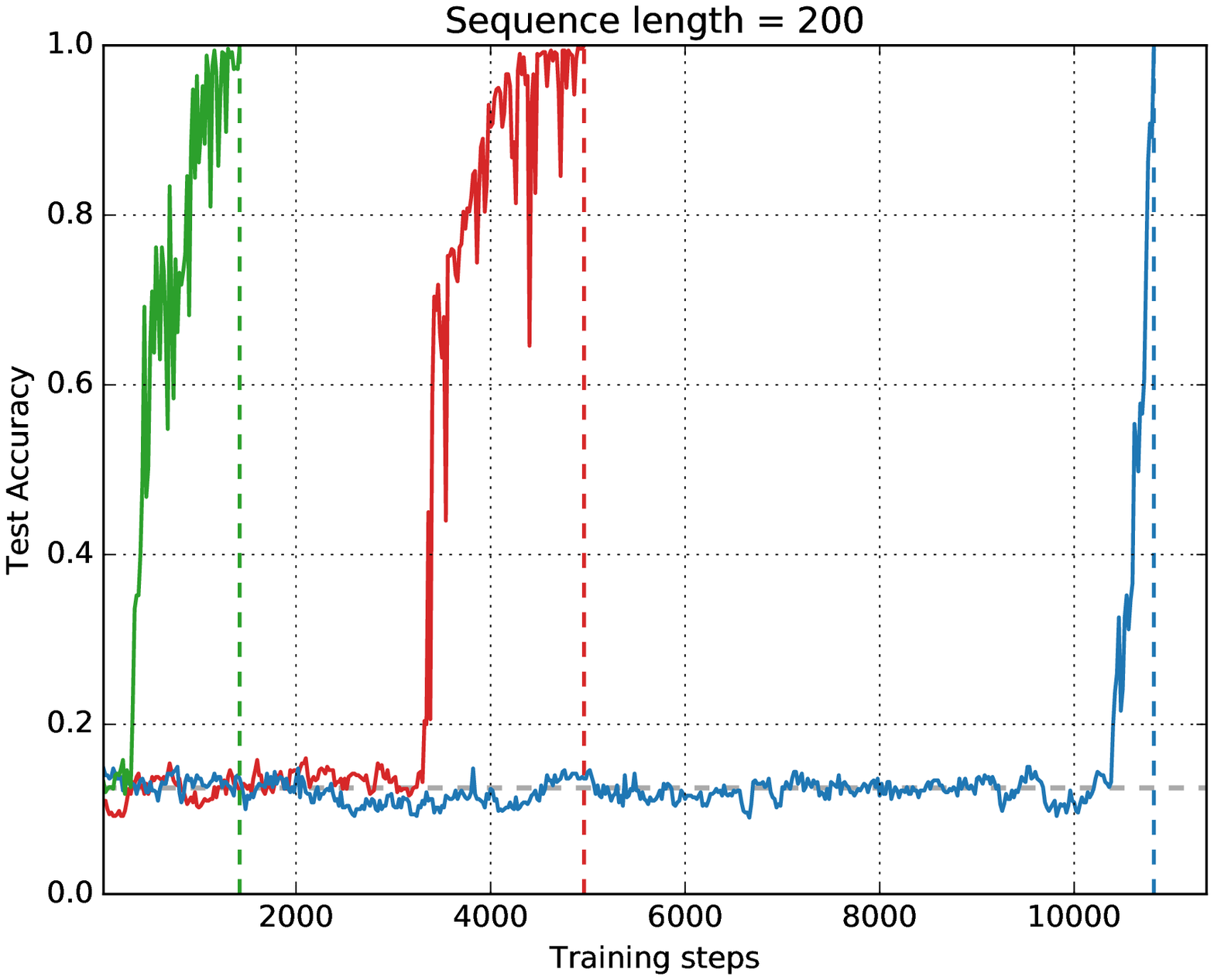}
  \includegraphics[width=0.49\textwidth,keepaspectratio]{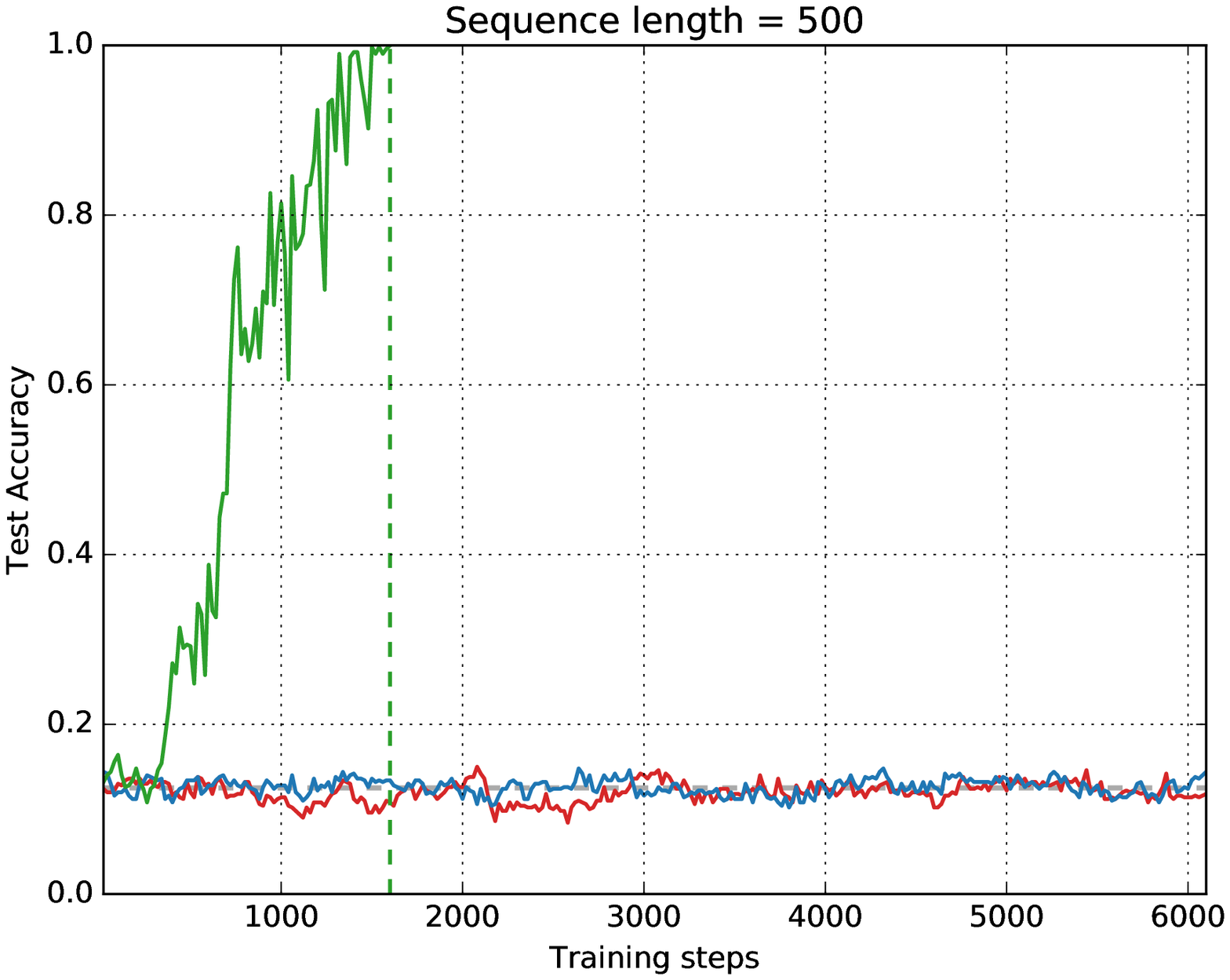}
  \includegraphics[width=0.49\textwidth,keepaspectratio]{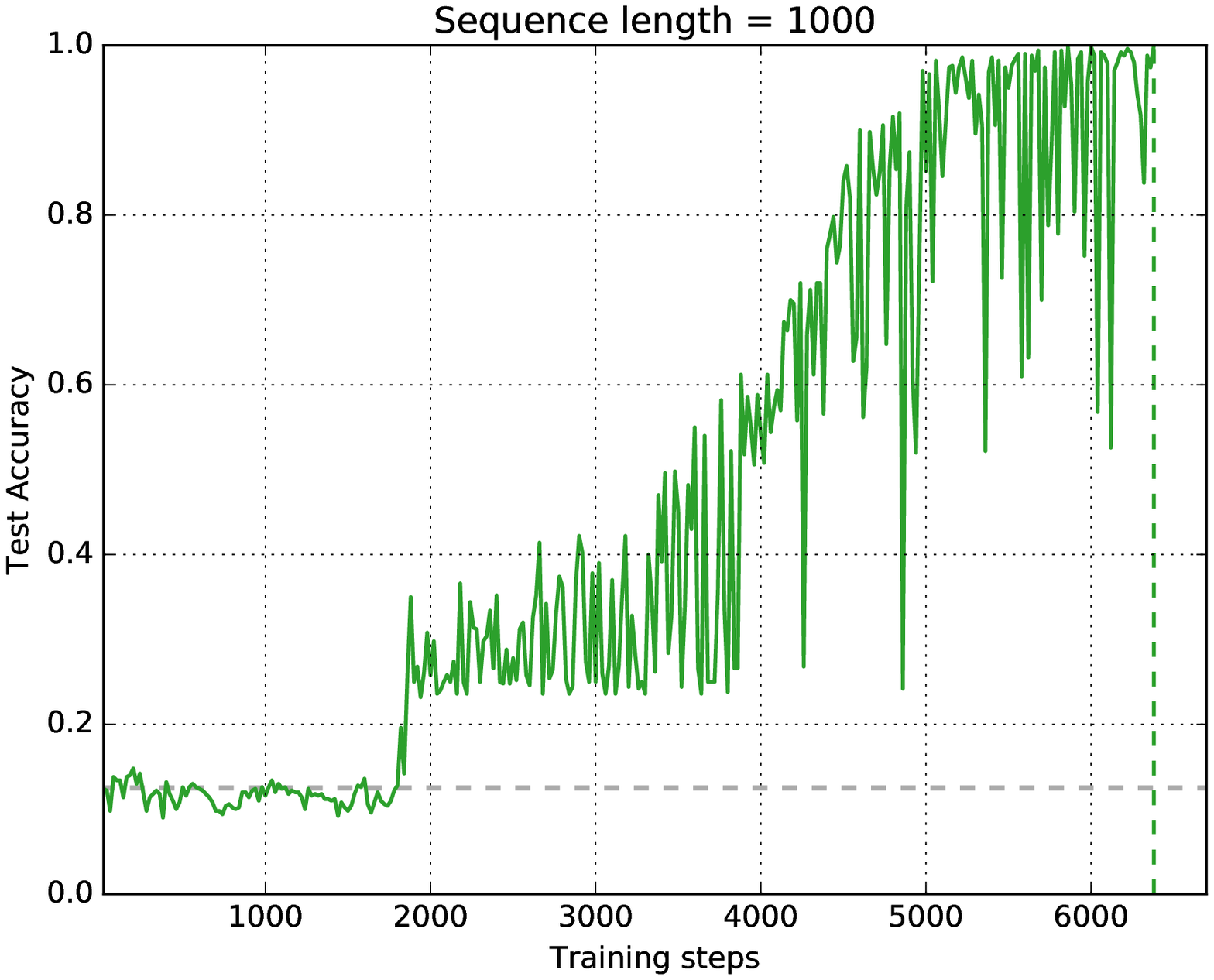}
  \caption{
    The results of the 3-bits temporal order problem on different sequence lengths. 
    The legends containing the information of model size are only shown in the first sub-figure.
    Each trial was stopped if all sequences in testing set are classified correctly,  as indicated by a dashed vertical line.
    When sequence length is 500, both LSTM and GRU failed within 50000 training steps,
    and their accuracy curves, which keeps fluctuating around the baseline are partially plotted.
  }
  \label{fig:to}
\end{figure}

The results are shown in Fig.\ref{fig:to}.
In contrast to the results of the adding problem, LSTM outperforms GRU on this task.
However, both LSTM and GRU fail in learning to distinguish the temporal order when the sequence length increases to $500$.
The GDU model with $\pu=0.1$ always starts learning earlier.
When trained with relatively longer sequences, GDU outperforms these $2$ models by a large margin with much less parameters.

\subsection{Multi-embedded Reber grammar}\label{sss:merg}

\begin{figure}[htpb]
  \centering
  \includegraphics[width=\textwidth,height=\textheight,keepaspectratio]{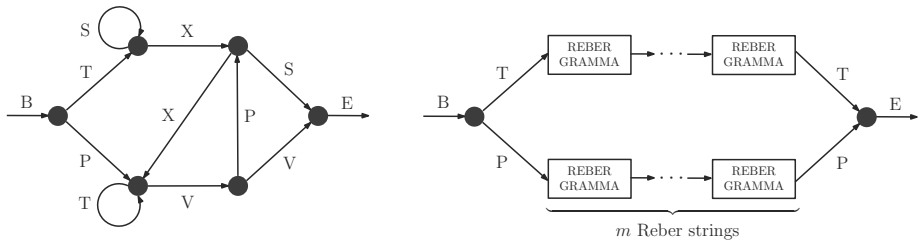} 
  \caption{
    \textbf{Left}: Trainsition diagram for the Reber grammar.
    \textbf{Right}: Transition diagram for the multi-embedded Reber grammar. Each box represents a Reber string. 
  }
  \label{merg}
\end{figure}

Embeded Reber grammar (ERG) \cite{fahlman91,lstm} is a good example containing dependencies with different time scales.
This task needs RNNs to read strings, one symbol at a time, and to predict the next symbol (error signals occur at every time step).
To correctly predict the symbol before last, a model has to remember the second symbol.
However, since it allows for training sequences with short time lags (of as few as $9$ steps), 
using it to evaluate a model's ability to learn long-term dependency is not appropriate.
In order to make the training sequences longer, we modified the ERG by having multiple Reber strings embedded between
the second and the last but one symbols (See Fig.\ref{merg}).

We refer to this variant as the \emph{multi-embedded Reber grammar (mERG)} and simply use the prefix $m$ to indicate the 
number of embedded Reber strings. 
For example, ``BT(BPVVE)(BTSXSE)(BTXXVVE)TE'' is a 3ERG sequence.
Since each Reber string has a minimal length $5$, the shortest $m$ERG sequence has a length of $5m+4$.

Learning $m$ERG requires a recurrent model to have the ability to latch long-term memory while keeping mid- and short-term 
memory (provided $m$ is big) in the meantime.
Further, there may be two legal successors from a given symbol and the model will never be able to do a perfect job of prediction.
During training, the rules defining the grammar are never presented. 
Thus the model will see contradictory examples, sometimes with one successor and sometimes the other, 
which requires it to learn to activate both legal outputs.
What's more, a model must remember how many Reber strings it has read to make a correct prediction of 
the next symbol if the current symbol is an E.
In other words, models must learn to \textbf{count}.

We set $m$ to be $10$, $20$ and $40$ for this task, with the minimal sequence length $54$, $104$ and $204$ respectively.
One sequence is given at a time.
As for the symbols with $2$ legal successors, a prediction is considered correct 
if the two desired outputs are the two with the largest values.
For each $m$ we generated $1000$ sequences for training and $256$ sequences for testing.
The sequences in testing set are unique and have never appeared in training set.
The same training and testing sets are used for comparing all models.

We also defined two criteria to test the model's ability to capture long- and short-term dependencies separately.
The one for short-term dependency is $SC$ (short for short-term criterion) defined as the 
percentages of testing sequences each symbol of which is predicted correctly by the model except for the one before last. 
The other is $LC$ (short for long-term criterion) defined as the 
percentages of testing sequences whose last but one symbol is predicted correctly.
We stopped the training when both $SC$ and $LC$ are satisfied (reach to $1$), 
namely all symbols in all testing sequences are predicted correctly. 
A naive strategy of predicting the symbol before last as T or P gives an expected $LC$ of $0.5$, which serves as the baseline.

An LSTM model and a GRU model both with 100 hidden states were chosen to be compared as previous,
with corresponding parameter numbers $43.9K$ and $33.1K$.
As for GDU, we chose a model with $35$ groups of size $2$ and $3$ groups of size $10$ (denoted as GDU(2x35+10x3)),
totally $100$ hidden units and $22.3K$ parameters.

\begin{figure}[htpb]
  \centering
  \includegraphics[width=0.49\textwidth,keepaspectratio]{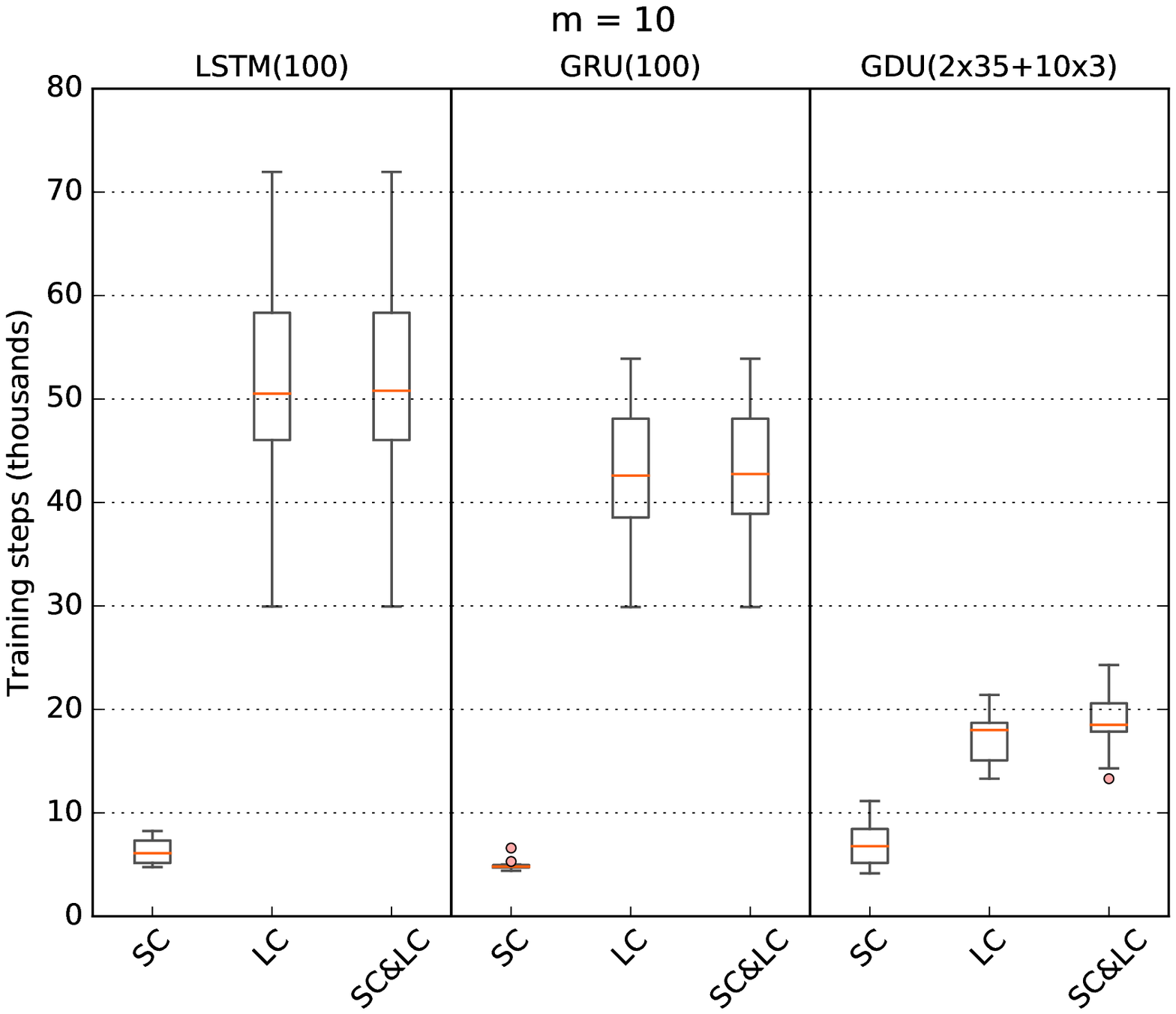}
  \includegraphics[width=0.49\textwidth,keepaspectratio]{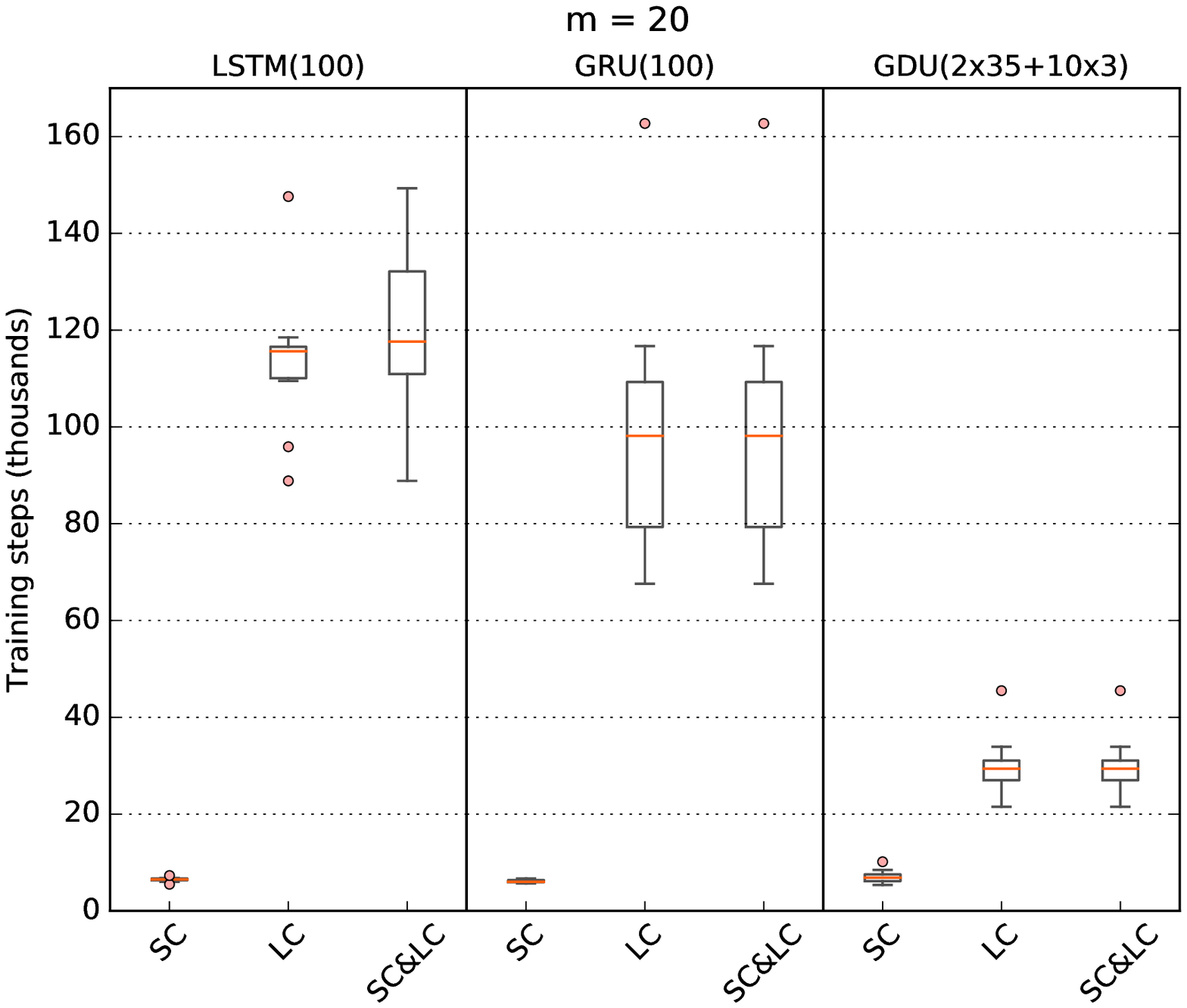}
  \includegraphics[width=0.49\textwidth,keepaspectratio]{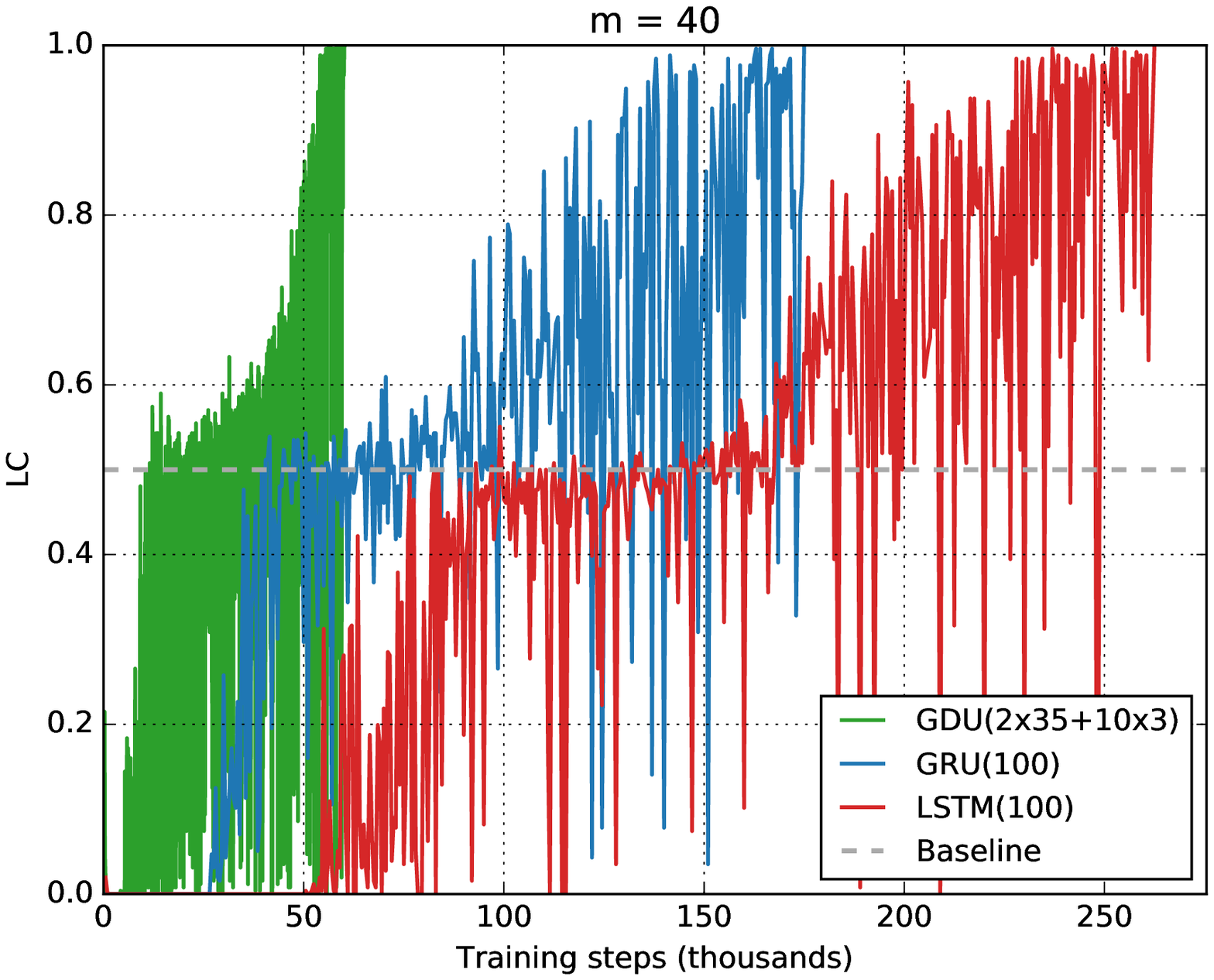}
  \includegraphics[width=0.49\textwidth,keepaspectratio]{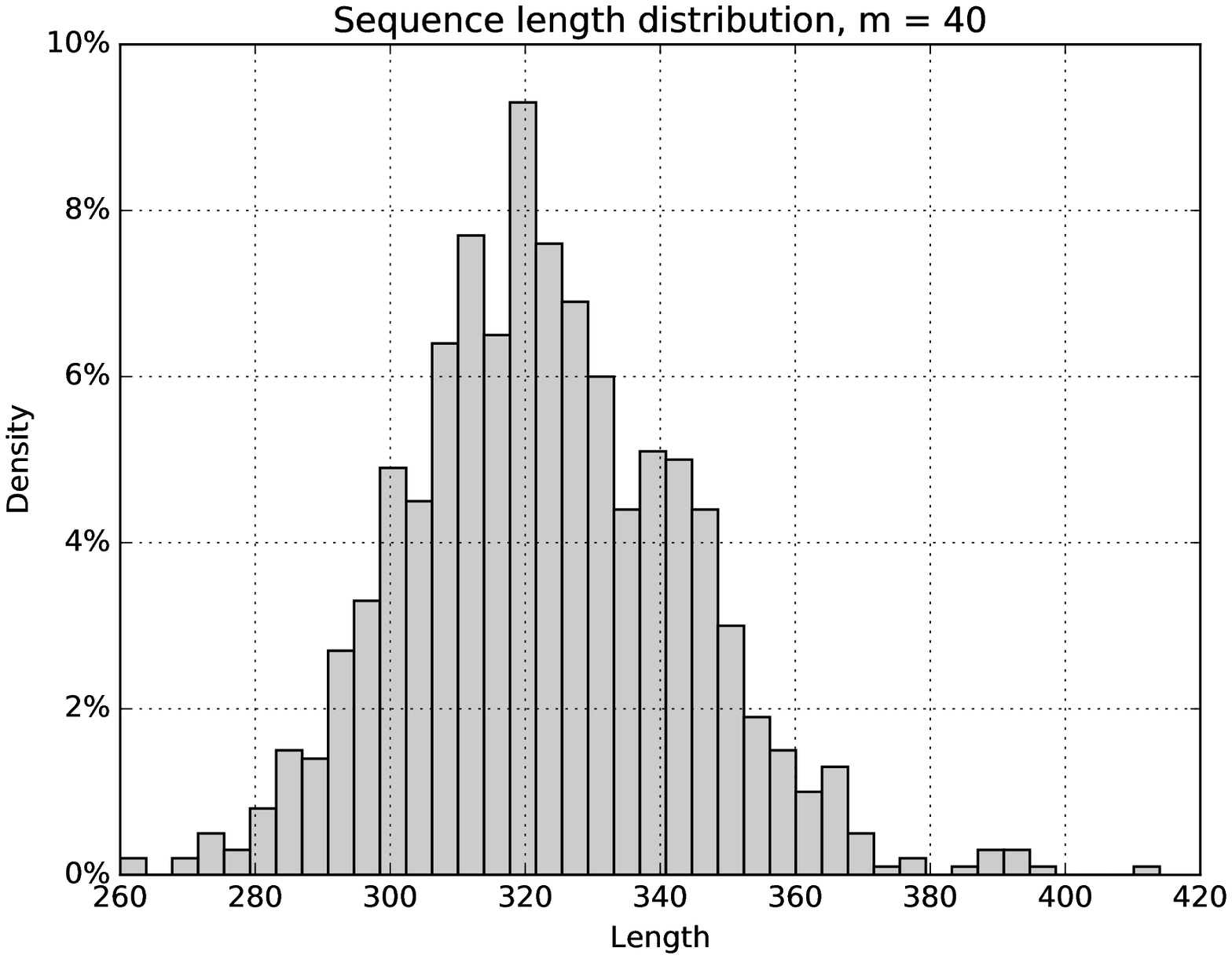}
  \caption{
    The results of the multi-embedded Reber grammar.
    The \textbf{upper left and right} figures show the training steps each model takes to satisfy the criteria (reach to 1.0) 
    for $m=10$ or $20$.
    Each box-whisker (showing median, $25\%$ and $75\%$ quantiles, minimum, maximum and outliers) contains the corresponding results
    of $10$ trials.
    For $m=40$ we only give the best results of each model in the \textbf{bottom left} figure.
    The \textbf{bottom right} figure shows the density histogram of sequence lengths in $40$ERG training set.
  }
  \label{fig:merg}
\end{figure}

From the results presented in Fig. \ref{fig:merg}, we can see for $m$ERG, models always learn to capture the short-term dependencies first.
While the long-term dependency is much more difficult to learn.
GRU outperforms LSTM this time, no matter from the aspect of which criterion.
GDU is slightly inferior to LSTM and GRU in terms of $SC$.
However, on aspect of $LC$, it has an obvious advantage. 

\begin{figure}[htpb]
  \centering
  \includegraphics[width=0.49\textwidth,keepaspectratio]{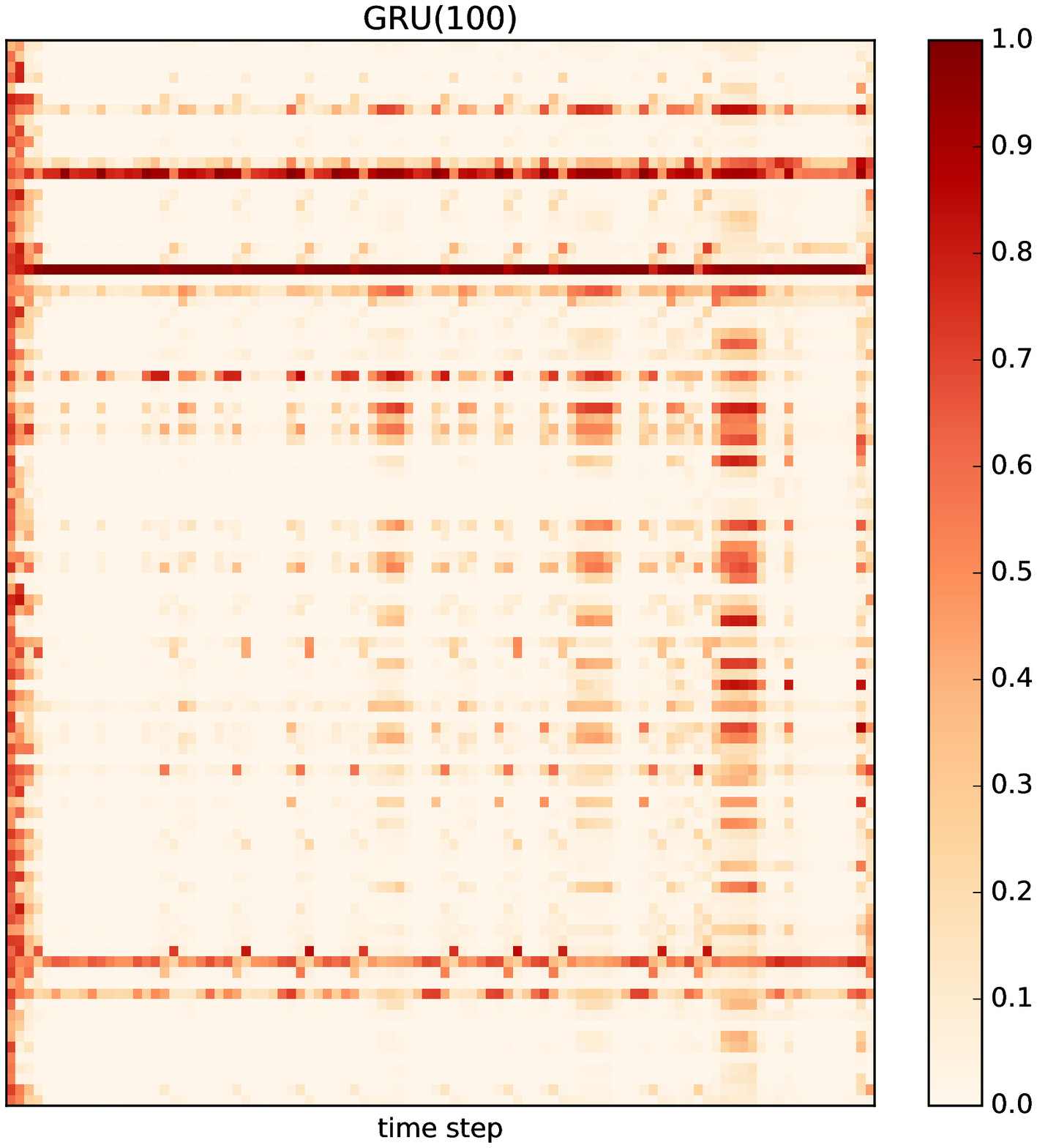}
  \includegraphics[width=0.49\textwidth,keepaspectratio]{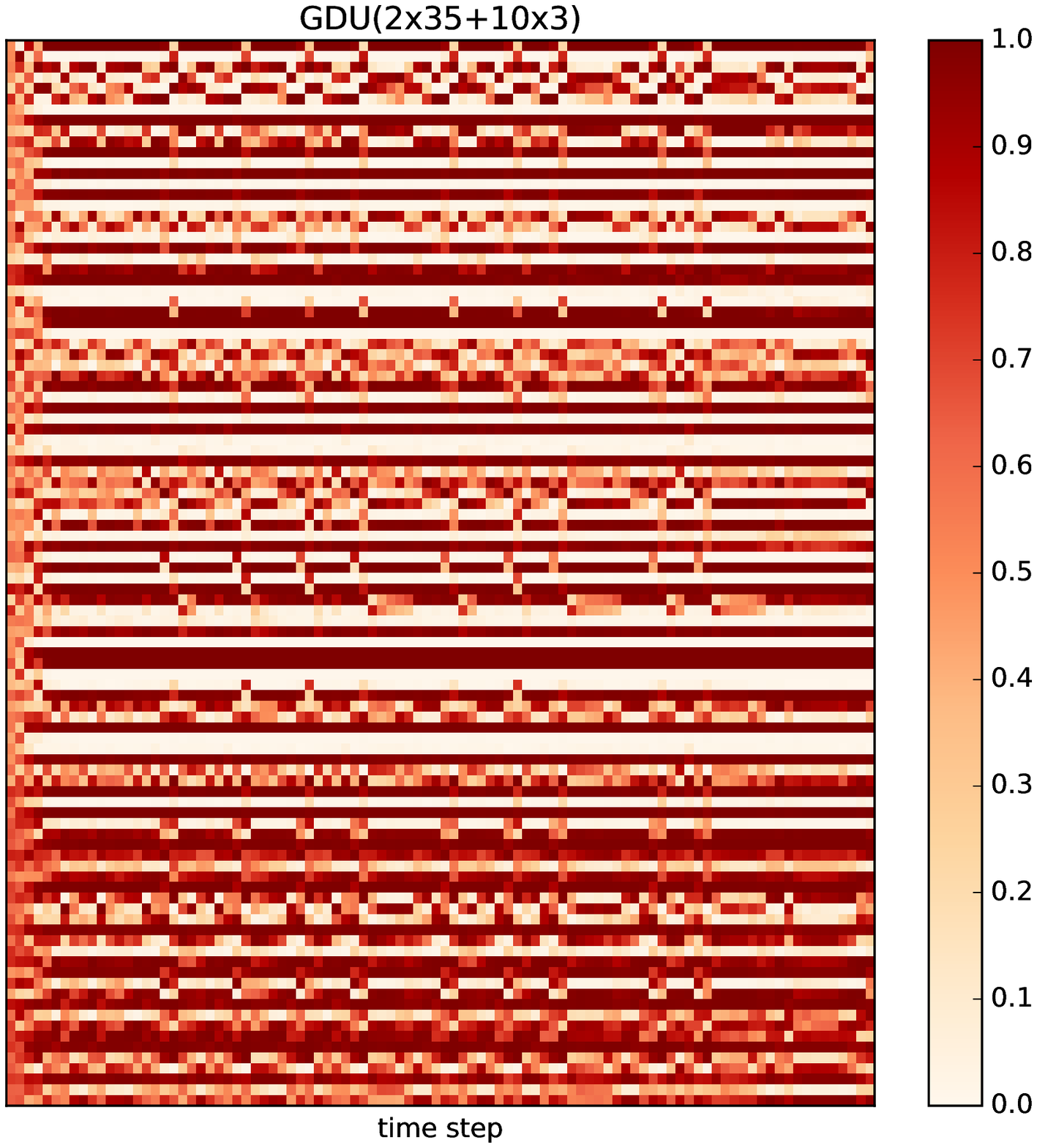}
  \caption{ 
    The activation of $\gv{\beta}$ of GRU(100) (\textbf{left}) and GDU(2x35+10x3) 
    (\textbf{right}) on a same sequence from 10ERG testing set.
    Each column corresponds to the gate activation at one time step. 
    Each row with continuous dark color corresponds to a gate unit which keeps active and thus latches information. 
  }
  \label{fig:gxu_on_10erg}
\end{figure}

As discussed in Section \ref{ss:bk}, learning to latch long-term information in the presence of short-term dependencies is
difficult for a traditional GAST model due to the gradient conflict.
GDU greatly alleviate this problem by limiting $\pu$ in cGAST, namely the proportion of states to be overwritten,
which results in a broader ``bandwidth'' for long-term information flow.
Fig. \ref{fig:gxu_on_10erg} illustrates this by visualizing the $\gv{\beta}$ activation of GAST models on a same $10$ERG sequence
after the $LC$ has been satisfied.

\subsection{Sequential pMNIST classification}

The sequential MNIST task \cite{irnn} can be seen as a sequence classification task in which $28\times28$ MNIST images \cite{lecun98}
of $10$ digits are read pixel by pixel from left to right, top to bottom.
While the sequential $p$MNIST \cite{irnn} is a challenging variant where the pixels are permuted by a same 
randomly generated permutation matrix.
This creates many longer term dependencies across pixels than in the original pixel ordering,
which makes it necessary for a model to learn and remember more complicated dependencies embedded in varying time scales.

\begin{table}
  \centering
  \begin{tabular}{lcc}
    \hline
    Model         & \# parameters ($\approx$, $K$) & Test Accuracy  \\
    \hline
    LSTM(128)     & 67.9                           & 91.2           \\
    GRU(128)      & 51.2                           & 90.6           \\
    GDU(4x32)     & 34.6                           & \textbf{93.5}  \\
    GDU(5x25)     & 33.0                           & 93.0           \\
    \hline
    LSTM(256)     & 266.8                          & 91.8           \\
    GRU(256)      & 200.7                          & 92.6           \\
    GDU(4x62)     & 134.7                          & 94.7           \\
    GDU(5x51)     & 133.6                          & \textbf{94.8}  \\
    \hline
  \end{tabular}
  \caption{ 
    Results for permuted pixel-by-pixel MNIST.
    Best result in each model set are bold.
  }
  \label{tab:pmnits}
\end{table}

All models are trained with batch size of $100$ and the learning rate is set to $0.001$.
No tricks, such as dropout \cite{dropout}, gradient clipping \cite{pascanu13}, recurrent batch normalization \cite{rbn16}, etc.,
are used since we are not focusing on achieving absolute high accuracy.
We trained two sets of models with $128$ and $256$ hidden states respectively. 
Again, GDU outperforms LSTM and GRU with less parameters in this task as shown in Table \ref{tab:pmnits}.

\begin{figure}[htpb]
  \centering
  \includegraphics[width=0.49\textwidth,keepaspectratio]{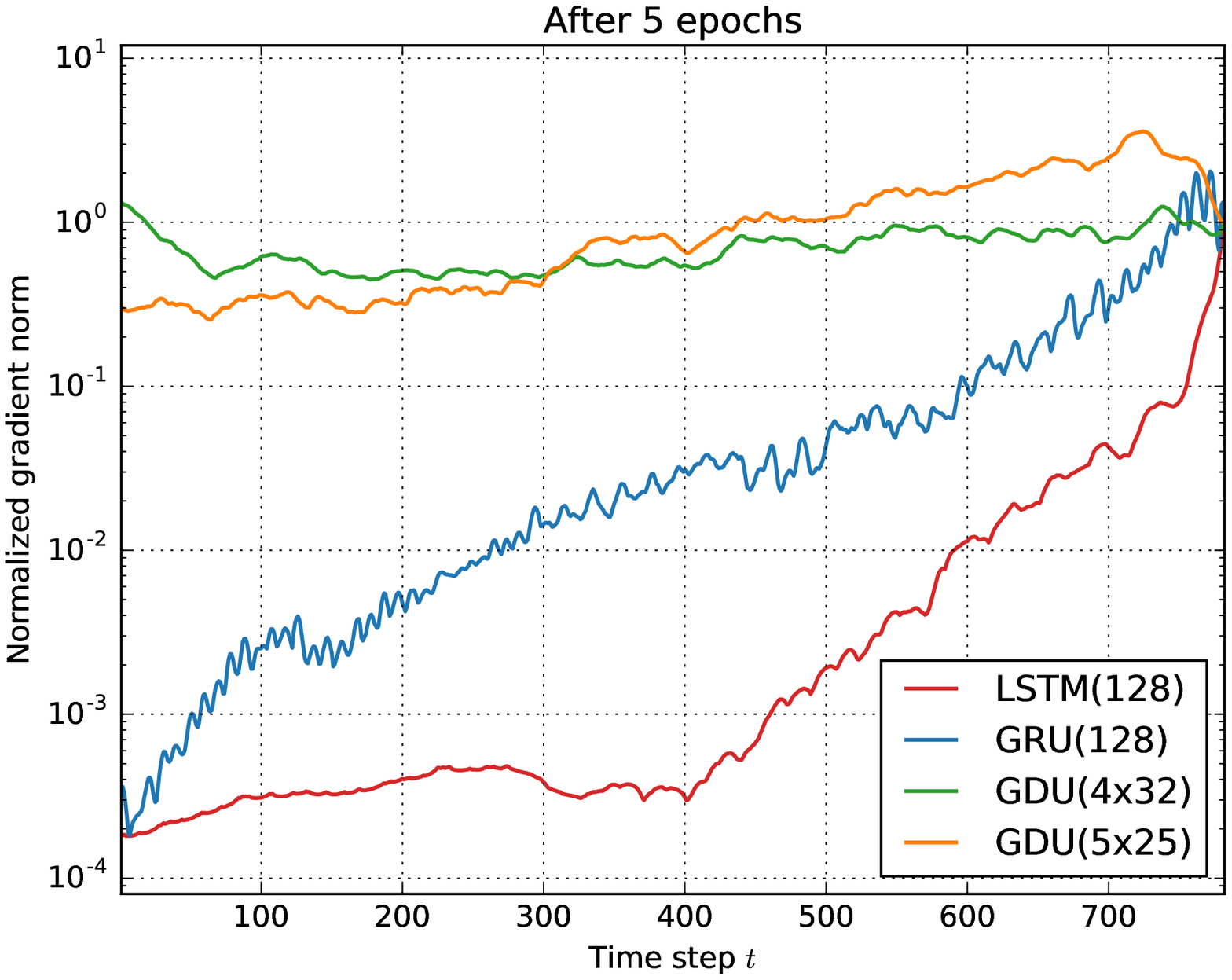}
  \includegraphics[width=0.49\textwidth,keepaspectratio]{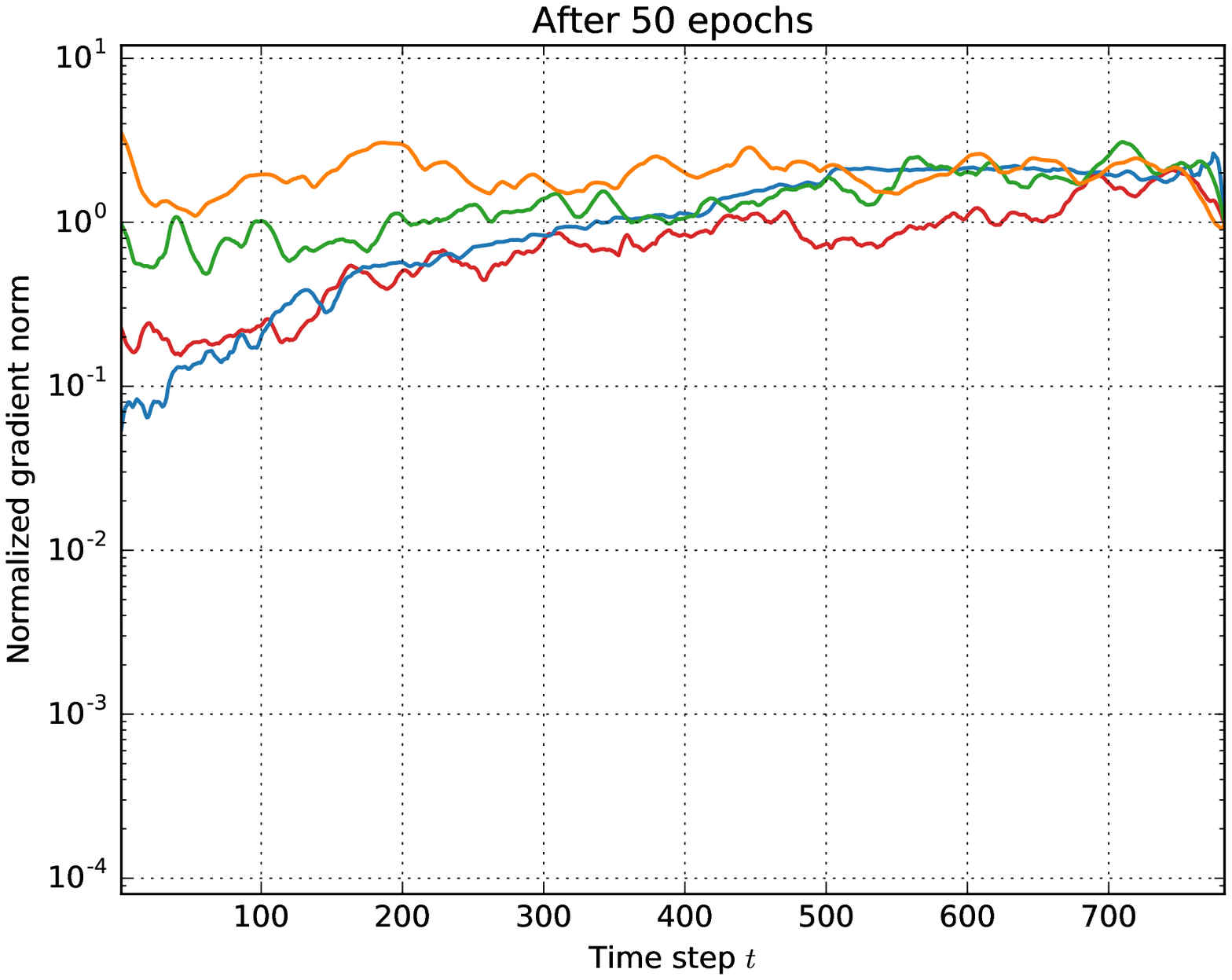}
  \caption{ 
    Norms of the error signal back-propagated to each time step, i.e. $\norm{\es{t}{784}} = \norm{\frac{\mathcal{L}}{\stt}}$ 
    after $5$ epochs (\textbf{left}) and $50$ epochs (\textbf{right}).
    For LSTM model, we calculate $\norm{\frac{\mathcal{L}}{\hat{\mathbf{s}}_t}}$ instead of $\norm{\frac{\mathcal{L}}{\stt}}$, 
    where $\hat{\mathbf{s}}_t$ is a concatenation of $\stt$ and $\mathbf{h}_t = \gamma_t(\stt)$.
  }
  \label{fig:merg_norm}
\end{figure}

As discussed in Section \ref{ss:bk}, controlling $\norm{\frac{\mathcal{L}}{\stt}}$ is the key to avoid the vanishing gradient issue, 
so that long-term dependencies can be learned.
We explored how each model propagated gradients by examining $\norm{\frac{\mathcal{L}}{\stt}}$ as a function of $t$, 
where $\mathcal{L}$ is the prediction loss.
Gradient norms were computed after $5$ and $50$ epochs and the normalized curves are plotted in Fig. \ref{fig:merg_norm}.
For LSTM and GRU, we can see that error signals have trouble in reaching far from where they are injected at the early stage.
This problem is reduced after training for dozens of epochs.
GDU models have better gradient properties than LSTM and GRU because of the distributor mechanism in Eqs. (\ref{eqn:sog}).

\section{Conclusions and future work}

We proposed a novel RNN architecture with gated additive state transition which contains only one gate unit.
The issues of gradient vanishing and conflict are mitigated by explicitly limiting the proportion of states to be overwritten 
at each time step.
Our experiments mainly focused on challenging pathological problems.
The results were consistent over different tasks and clearly demonstrated that the proposed grouped distributor architecture
is helpful to extract long-term dependencies embedded in data.

A plethora of further ideas can be explored based on our findings.
For example, various combinations of groups with different sizes and overwrite proportions can be explored.
Further, the overwrite proportion $\delta$ can be trained.
What's more interesting is that the grouped distributor structure can be used spatially to ease gradient-based training 
of very deep networks.
To be more specific, this work can base on the \emph{highway network} \cite{highway} 
in which the distributor operator can be used to calculate the \emph{transform gate}.
Testings of the stacked GDU on other data sets are also planned.

\bibliographystyle{unsrt}
\bibliography{gdu}

\begin{thebibliography}{10}

\bibitem{rumelhart86}
David~E. Rumelhart, Geoffrey~E. Hinton, and Ronald~J. Williams.
\newblock Learning representations by back-propagating errors.
\newblock {\em Nature}, 323:533--536, 1986.

\bibitem{werbos88}
Paul J.Werbos.
\newblock Generalization of backpropagation with application to a recurrent gas
  market model.
\newblock {\em Neural Networks}, 1:339--356, 1988.

\bibitem{hochreiter01}
Sepp Hochreiter, Yoshua Bengio, and Paolo Frasconi.
\newblock Gradient flow in recurrent nets: the difficulty of learning long-term
  dependencies.
\newblock In J.~Kolen and S.~Kremer, editors, {\em Field Guide to Dynamical
  Recurrent Networks}. IEEE Press, 2001.

\bibitem{lstm}
Sepp Hochreiter and J\"{u}rgen Schmidhuber.
\newblock Long short-term memory.
\newblock {\em Neural Comput.}, 9(8):1735--1780, 1997.

\bibitem{martens10}
James Martens.
\newblock Deep learning via hessian-free optimization.
\newblock In {\em Proceedings of the 27th International Conference on
  International Conference on Machine Learning}, ICML'10, pages 735--742, USA,
  2010. Omnipress.

\bibitem{martens11}
James Martens and Ilya Sutskever.
\newblock Learning recurrent neural networks with hessian-free optimization.
\newblock In {\em Proceedings of the 28th International Conference on
  International Conference on Machine Learning}, ICML'11, pages 1033--1040,
  USA, 2011. Omnipress.

\bibitem{saxe13}
Andrew~M. Saxe, James~L. McClelland, and Surya Ganguli.
\newblock Exact solutions to the nonlinear dynamics of learning in deep linear
  neural networks, 2013.

\bibitem{irnn}
Quoc~V. Le, Navdeep Jaitly, and Geoffrey~E. Hinton.
\newblock A simple way to initialize recurrent networks of rectified linear
  units, 2015.

\bibitem{narx}
Tsungnan Lin, B.~G. Horne, P.~Tino, and C.~L. Giles.
\newblock Learning long-term dependencies in narx recurrent neural networks.
\newblock {\em Trans. Neur. Netw.}, 7(6):1329--1338, 1996.

\bibitem{dilatedrnn}
Shiyu Chang, Yang Zhang, Wei Han, Mo~Yu, Xiaoxiao Guo, Wei Tan, Xiaodong Cui,
  Michael Witbrock, Mark~A Hasegawa-Johnson, and Thomas~S Huang.
\newblock Dilated recurrent neural networks.
\newblock In I.~Guyon, U.~V. Luxburg, S.~Bengio, H.~Wallach, R.~Fergus,
  S.~Vishwanathan, and R.~Garnett, editors, {\em Advances in Neural Information
  Processing Systems 30}, pages 77--87. Curran Associates, Inc., 2017.

\bibitem{mistrnn}
Robert DiPietro, Christian Rupprecht, Nassir Navab, and Gregory~D. Hager.
\newblock Analyzing and exploiting {NARX} recurrent neural networks for
  long-term dependencies, 2018.

\bibitem{gru}
Kyunghyun Cho, Bart van Merrienboer, Caglar Gulcehre, Dzmitry Bahdanau, Fethi
  Bougares, Holger Schwenk, and Yoshua Bengio.
\newblock Learning phrase representations using {RNN} encoder-decoder for
  statistical machine translation, 2014.

\bibitem{ugrnn}
David~Sussillo Jasmine~Collins, Jascha Sohl-Dickstein.
\newblock Capacity and trainability in recurrent neural networks.
\newblock In {\em International Conference on Learning Representations}, 2016.

\bibitem{mgu}
Guo-Bing Zhou, Jianxin Wu, Chen-Lin Zhang, and Zhi-Hua Zhou.
\newblock Minimal gated unit for recurrent neural networks, 2016.

\bibitem{cwrnn}
Jan Koutnik, Klaus Greff, Faustino Gomez, and Juergen Schmidhuber.
\newblock A clockwork {RNN}.
\newblock In Eric~P. Xing and Tony Jebara, editors, {\em Proceedings of the
  31st International Conference on Machine Learning}, volume~32 of {\em
  Proceedings of Machine Learning Research}, pages 1863--1871, Bejing, China,
  2014. PMLR.

\bibitem{elman90}
Jeffrey~L. Elman.
\newblock Finding structure in time.
\newblock {\em COGNITIVE SCIENCE}, 14(2):179--211, 1990.

\bibitem{pascanu13}
Razvan Pascanu, Tomas Mikolov, and Yoshua Bengio.
\newblock On the difficulty of training recurrent neural networks.
\newblock In {\em Proceedings of the 30th International Conference on
  International Conference on Machine Learning - Volume 28}, ICML'13, pages
  III--1310--III--1318. JMLR.org, 2013.

\bibitem{forgetgate}
Felix~A. Gers, J\"{u}rgen~A. Schmidhuber, and Fred~A. Cummins.
\newblock Learning to forget: Continual prediction with lstm.
\newblock {\em Neural Comput.}, 12(10):2451--2471, October 2000.

\bibitem{peephole}
Felix~A. Gers and Juergen Schmidhuber.
\newblock Recurrent nets that time and count.
\newblock Technical report, 2000.

\bibitem{greff17}
Klaus Greff; Rupesh K. Srivastava; Jan Koutn\'{\i}k ; Bas R. Steunebrink
  ;~J\"{u}rgen Schmidhuber.
\newblock Lstm: A search space odyssey.
\newblock {\em IEEE Transactions on Neural Networks and Learning Systems},
  28(8):2222--2232, 2017.

\bibitem{chung14}
Junyoung Chung, Caglar Gulcehre, Kyunghyun Cho, and Yoshua Bengio.
\newblock Empirical evaluation of gated recurrent neural networks on sequence
  modeling.
\newblock In {\em NIPS 2014 Workshop on Deep Learning, December 2014}, 2014.

\bibitem{odyssey}
Klaus Greff, Rupesh~Kumar Srivastava, Jan Koutník, Bas~R. Steunebrink, and
  Jürgen Schmidhuber.
\newblock Lstm: A search space odyssey.
\newblock {\em CoRR}, abs/1503.04069, 2015.

\bibitem{hihi95}
Salah El~Hihi and Yoshua Bengio.
\newblock Hierarchical recurrent neural networks for long-term dependencies.
\newblock 1996.

\bibitem{glorot10}
Xavier Glorot and Yoshua Bengio.
\newblock Understanding the difficulty of training deep feedforward neural
  networks.
\newblock In {\em In Proceedings of the International Conference on Artificial
  Intelligence and Statistics (AISTATS’10). Society for Artificial
  Intelligence and Statistics}, 2010.

\bibitem{adam}
Diederik~P. Kingma and Jimmy Ba.
\newblock Adam: {A} method for stochastic optimization.
\newblock In {\em {ICLR}}, 2015.

\bibitem{tf}
Martín Abadi, Ashish Agarwal, Paul Barham, Eugene Brevdo, Zhifeng Chen, Craig
  Citro, Greg Corrado, Andy Davis, Jeffrey Dean, Matthieu Devin, Sanjay
  Ghemawat, Ian Goodfellow, Andrew Harp, Geoffrey Irving, Michael Isard,
  Yangqing Jia, Rafal Jozefowicz, Lukasz Kaiser, Manjunath Kudlur, Josh
  Levenberg, Dan Mané, Rajat Monga, Sherry Moore, Derek Murray, Chris Olah,
  Mike Schuster, Jonathon Shlens, Benoit Steiner, Ilya Sutskever, Kunal Talwar,
  Paul Tucker, Vincent Vanhoucke, Vijay Vasudevan, Fernanda Viégas, Oriol
  Vinyals, Pete Warden, Martin Wattenberg, Martin Wicke, Yuan Yu, and Xiaoqiang
  Zheng.
\newblock Tensorflow: Large-scale machine learning on heterogeneous distributed
  systems, 2015.

\bibitem{urnn}
Martin Arjovsky, Amar Shah, and Yoshua Bengio.
\newblock Unitary evolution recurrent neural networks, 2015.

\bibitem{fahlman91}
Scott~E. Fahlman.
\newblock The recurrent cascade-correlation architecture.
\newblock In R.~P. Lippmann, J.~E. Moody, and D.~S. Touretzky, editors, {\em
  Advances in Neural Information Processing Systems 3}, pages 190--196.
  Morgan-Kaufmann, 1991.

\bibitem{lecun98}
Yann Lecun, Léon Bottou, Yoshua Bengio, and Patrick Haffner.
\newblock Gradient-based learning applied to document recognition.
\newblock In {\em Proceedings of the IEEE}, pages 2278--2324, 1998.

\bibitem{dropout}
Nitish Srivastava, Geoffrey Hinton, Alex Krizhevsky, Ilya Sutskever, and Ruslan
  Salakhutdinov.
\newblock Dropout: A simple way to prevent neural networks from overfitting.
\newblock {\em Journal of Machine Learning Research}, 15:1929--1958, 2014.

\bibitem{rbn16}
Tim Cooijmans, Nicolas Ballas, C{\'{e}}sar Laurent, and Aaron~C. Courville.
\newblock Recurrent batch normalization, 2016.

\bibitem{highway}
Rupesh~Kumar Srivastava, Klaus Greff, and J\"{u}rgen Schmidhuber.
\newblock Highway networks, 2015.

\end{thebibliography}

\newpage

\end{document}